\documentclass{article}

\usepackage{arxiv}

\usepackage{graphicx}  
\usepackage{amsmath,bm}
\usepackage{amssymb,booktabs}
\usepackage{enumitem}
\usepackage{algorithm}
\usepackage{algorithmic}
\usepackage{caption}
\usepackage[sort&compress,numbers]{natbib}
\usepackage{bm}
\usepackage{multirow}
\usepackage{xspace,soul}
\usepackage{url}
\usepackage[para,online,flushleft]{threeparttable}
\usepackage{mathtools}

\usepackage{caption}
\usepackage{tikz}
\usepackage{verbatim}
\usepackage[noadjust]{cite}
\usepackage{enumerate}
\usetikzlibrary{fit,shapes.geometric,backgrounds,positioning-plus,calc,matrix}

\newcommand{\method}{\mbox{$\mathop{\mathtt{CDC}}\limits$}\xspace}
\newcommand{\methodD}{\mbox{$\mathop{\mathtt{CDC\!\texttt{+}}}\limits$}\xspace}

\newcommand{\relation}{\mbox{$\mathop{\mathtt{r}}\limits$}\xspace}
\newcommand{\head}{\mbox{$\mathop{\mathtt{h}}\limits$}\xspace}
\newcommand{\tail}{\mbox{$\mathop{\mathtt{t}}\limits$}\xspace}
\newcommand{\entity}{\mbox{$\mathop{\mathtt{e}}\limits$}\xspace}

\newcommand{\newhead}{\mbox{$\mathop{\mathtt{N\text{-}h}}\limits$}\xspace}
\newcommand{\newtail}{\mbox{$\mathop{\mathtt{N\text{-}t}}\limits$}\xspace}
\newcommand{\newht}{\mbox{$\mathop{\mathtt{N\text{-}ht}}\limits$}\xspace}

\newcommand{\m}{\mbox{$\mathop{\mathtt{m}}\limits$}\xspace}

\newcommand{\tSet}{\mbox{$\mathop{\mathrm{S}}\limits$}\xspace}
\newcommand{\eSet}{\mbox{$\mathop{\mathrm{E}}\limits$}\xspace}
\newcommand{\rSet}{\mbox{$\mathop{\mathrm{R}}\limits$}\xspace}

\newcommand{\e}{\mbox{$\mathop{\mathbf{v}}\limits$}\xspace}
\newcommand{\etxt}{\mbox{$\mathop{\mathbf{v}^d}\limits$}\xspace}

\newcommand{\fflata}{\mbox{$\mathop{\mathbf{f}}\limits$}\xspace}
\newcommand{\fflatb}{\mbox{$\mathop{\mathbf{f}^s}\limits$}\xspace}

\newcommand{\fflattxt}{\mbox{$\mathop{\mathbf{f}^d}\limits$}\xspace}
\newcommand{\fflatstr}{\mbox{$\mathop{\mathbf{f}^u}\limits$}\xspace}
\newcommand{\ffcltxt}{\mbox{$\mathop{\mathbf{g}^d}\limits$}\xspace}
\newcommand{\ffclstr}{\mbox{$\mathop{\mathbf{g}^u}\limits$}\xspace}

\newcommand{\ffcla}{\mbox{$\mathop{\mathbf{g}}\limits$}\xspace}
\newcommand{\ffclb}{\mbox{$\mathop{\mathbf{g}^s}\limits$}\xspace}
\newcommand{\bfcl}{\mbox{$\mathop{\mathbf{b}_f}\limits$}\xspace}
\newcommand{\Wfcl}{\mbox{$\mathop{\mathrm{W}_f}\limits$}\xspace}
\newcommand{\blogit}{\mbox{$\mathop{\mathbf{b}_l}\limits$}\xspace}
\newcommand{\Wlogit}{\mbox{$\mathop{\mathrm{W}_l}\limits$}\xspace}

\newcommand{\etal}{\mbox{\emph{et al.}}\xspace}

\newcommand{\score}{\mbox{$\mathop{\mathrm{s}}\limits$}\xspace}
\newcommand{\scorea}{\mbox{$\mathop{\mathrm{s}}\limits$}\xspace}
\newcommand{\scoreb}{\mbox{$\mathop{\mathrm{s}^s}\limits$}\xspace}
\newcommand{\scorestr}{\mbox{$\mathop{\mathrm{s_{str}}}\limits$}\xspace}
\newcommand{\scoretxt}{\mbox{$\mathop{\mathrm{s_{txt}}}\limits$}\xspace}

\newcommand{\MRank}{\mbox{$\mathop{\mathtt{{MR}}}\limits$}\xspace}

\newcommand{\HITten}{\mbox{$\mathop{\mathtt{H@10}}\limits$}\xspace}

\graphicspath{{./}{./}}

\title{CNN-based Dual-Chain Models for
\\Knowledge Graph Learning}

\author{Bo Peng\\         
The Ohio State University\\
peng.707@buckeyemail.osu.edu\\
\And
Renqiang Min\\
NEC Laboratories America, Inc\\
renqiang@nec-labs.com\\
\And
Xia Ning\\
The Ohio State University\\
ning.104@osu.edu
}

\begin{document}

\maketitle

\begin{abstract}

  %
  Knowledge graph learning plays a critical role in integrating domain-specific 
  knowledge bases when deploying machine learning and data mining models in practice. 
  Existing methods on knowledge graph learning primarily focus on modeling
  the relations among entities as translations among the relations and entities, 
  and many of these methods are not able to handle zero-shot problems, when
  new entities emerge. 
  In this paper, we present a new convolutional neural network (CNN)-based
  dual-chain model.
  Different from translation based methods, in our model,
  interactions among relations and entities are directly captured via CNN
  over their embeddings. Moreover, 
  a secondary chain of learning is conducted simultaneously to incorporate additional
  information and to enable better performance.
  We also present an extension of this model, which incorporates descriptions
  of entities and learns a second set of entity embeddings from the descriptions.
  As a result, the extended model is able to effectively handle zero-shot problems.
  We conducted comprehensive experiments, comparing our methods with 15
  methods on 8 benchmark datasets.
  Extensive experimental results demonstrate that our proposed methods achieve or 
  outperform the state-of-the-art results on knowledge graph learning, and outperform
  other methods on zero-shot problems. In addition, our methods applied to real-world 
  biomedical data are able to produce results that conform to expert domain knowledge. 
\end{abstract}

\section{Introduction}
\label{sec:intro}
%
%
%
%
Real-world knowledge can be represented in directed, multi-relational, and structured
graphs, the so-called knowledge graphs.
In a knowledge graph, graph nodes represent entities of interest and edges encode
relations among entities. Therefore, one piece of knowledge can be succinctly represented
by two nodes and one edge between them in the knowledge graph,
where the source node is referred to \emph{head} and the destination node is
referred to \emph{tail}. Thus, the whole collection of such (head, relation, tail)
triplets captures entirely the knowledge graph structure and content. 
Real-world knowledge graphs include WordNet \citep{wordnet}, a large lexical database
for English, Google Knowledge Graph~\footnote{\url{https://developers.google.com/knowledge-graph/}}, 
a comprehensive system about facts among people, etc.,  
and DBpedia \citep{dbpedia}, a structured knowledge base for information extracted
from Wikipedia, etc.

Typically, knowledge graphs suffer from incompleteness, that is, many relations (i.e., edges)
among entities (i.e., nodes) in the graphs cannot be established due to the limited or unknown
information.
It is substantially significant to enable complete knowledge graphs. For example, a complete
knowledge graph on the relations between diseases and genes, and the relations between genes and
drugs, could dramatically help improve medicine and health care. 
However, knowledge graph completion, and in general, knowledge graph learning, is
a very challenging task, requiring strong modeling power to capture the underlying logic
that governs the formation of the head-relation-tail triplets, and thus, the knowledge.

Substantial research efforts have been dedicated to knowledge graph learning, in
particular, from the deep learning community, due to the extraordinary capability of
deep learning in modeling complicated relations among data. 
Quite a few deep learning methods~\citep{TransE, TransH, TransR, TransD, TranSparse, DistMult, ComplEx, ConvE}
(Section~\ref{sec:related}) have been developed and have dramatically improved knowledge graph
learning.
Still, there exist increasing needs to further improve knowledge graph learning performance, particularly
for high-impact applications such as medicine and health care.
In addition, being able to leverage additional information from other sources, for example, descriptions
on entities, to improve knowledge graph learning, is very promising but less explored. 

In this paper, we present our efforts toward improving knowledge graph learning as follows:
\begin{itemize}[noitemsep,nolistsep,leftmargin=*]
\item We present new deep learning methods, termed as
  \underline{C}onvolutional Neural Network (CNN)-based
  \underline{D}ual-\underline{C}hain methods, denoted as \method, for knowledge graph learning. 
%
  Very different from existing methods (Section~\ref{sec:related}), 
  \method models the interactions among relations and entities via a CNN over their respective
  embeddings as features for the corresponding triplet.  
  Then, logistic regression is used to predict whether the triplets are valid from the learned
  features.
\item In addition to the above primary learning process,
  we introduced a secondary chain of simultaneous learning process in \method,
  which learns from sparsified
  entity and relation embeddings in a same architecture as in the primary chain.
  This dual-chain structure in \method helps reduce overfitting, accelerate
  convergence and improve learning performance in general. 
\item We also developed an extension of \method, denoted as \methodD, 
  to incorporate and utilize textual information (i.e., entity descriptions) in
  knowledge graph learning.
  In \methodD, entities and relations will both have a second embedding learned
  from the descriptions via structure attention and CNNs. 
  %
  The embeddings learned from graph structures (i.e., the head-relation-tail triplets)
  and from the descriptions are used together along the dual chains to enable more accurate prediction on triplets.
  %
  Thus, \methodD is able to improve learning performance and handle zero-shot
  problems~\citep{DKRL}.
\item We conducted comprehensive experiments on 8 public, benchmark datasets, and compared
  our methods with another 15
methods on knowledge graph learning. 
  The experimental results demonstrate that our methods are able to achieve or outperform the
  state of the art. 
\item 
  We also investigated our methods on two important biological knowledge graphs:
  drug-gene interaction graph and drug-drug interaction graph.
  Our experimental results demonstrate that our methods outperform the state-of-the-art knowledge
  graph learning methods, and make discoveries supported by expert domain knowledge.
\end{itemize}

Table~\ref{tbl:notation} presents the key definitions and notations used in this paper.
The rest of this paper is organized as follows. Section~\ref{sec:related} presents the related
work. 
Section~\ref{sec:method} presents the details of \method method.
Section~\ref{sec:methodd} presents the details of \methodD method.
Section~\ref{sec:comp} compares \method with existing methods. 
Section~\ref{sec:exp} presents the experimental protocols, datasets and evaluation metrics.
Section~\ref{sec:results} presents the experimental results.
In Section~\ref{sec:conclusion}, we conclude our work and discuss future work.

\begin{table}[!h]
  \caption{Definitions and Notations}
  \label{tbl:notation}
  \centering
  \begin{threeparttable}
      \begin{tabular}{
	@{\hspace{2pt}}l@{\hspace{2pt}}
	@{\hspace{2pt}}l@{\hspace{2pt}}          
	}
        \toprule
        notation & meaning \\
        \midrule
        \entity/\eSet   & entity/entity set \\
        \relation/\rSet & relation/relation set \\
        \head/\tail  & head/tail entity \\
        (\head, \relation, \tail) & a head-relation-tail triplet \\
        \tSet/\tSet'        & set of positive/negative $(\head, \relation, \tail)$ triplets \\
        \e/$\e^{s}$          & embedding/sparsified embedding of size $k$\\
        \bottomrule
      \end{tabular}
  \end{threeparttable}
\end{table}
%


\vspace{-20pt}
\section{Related Work}
\label{sec:related}

In this section, we review the literature on knowledge graph learning,
particularly using deep neural networks, 
based on how the relations among entities are formulated and learned, and which information
is used to learn the relations.

\subsection{\mbox{Translation based Knowledge Graph Learning}}
\label{sec:related:trans}

The most popular way to formulate the relations among entities in knowledge
graph learning is through translation, pioneered by the work
of Bordes \etal~\citep{TransE}
on the TransE method. 
%
In TransE, a true relation \relation between head \head and tail \tail
is represented as a linear translation
among their respective embeddings $\e_{\scriptsize{\relation}}$, $\e_{\scriptsize{\head}}$ and
$\e_{\scriptsize{\tail}}$, that is, $\e_{\scriptsize{\head}} + \e_{\scriptsize{\relation}} = \e_{\scriptsize{\tail}}$.
Thus, a false relation $\relation^\prime$ between head $\head^\prime$ and tail $\tail^\prime$ will
induce a large distance between $\e_{\scriptsize{\head}} + \e_{\scriptsize{\relation}}$ and
$\e_{\scriptsize{\tail}}$.
All the entity and relation embeddings are learned via optimizing a pairwise ranking
objective among all the triplets, that is, distances from positive triplets are smaller than
the distances from their corresponding negative triplets.
Wang \etal~\citep{TransH} developed TransH,
which also uses the linear translation as in TransE to model relations. 
Different from TransE, in TransH, one hyperplane is learned for each relation.
Head and tail entities are projected onto the hyperplane of a relation as
relation-specific embeddings. The linear translation is done on the hyperplane
to predict if the relation holds for the projected heads and tails. 
TransR~\citep{TransR} is another translation based method. In TransR,
entities and relations are represented in two different spaces.
For each relation, a mapping function is learned to map the head and
tail entities from the entity space to the relation space, where
the relations between the head and tail entities are represented via the
translation as in TransE. 
Cluster-based TransR (CTransR)~\citep{TransR} is an extension of TransR.
In CTransR, the triplets of a same relation are clustered into groups.
In each group, a relation embedding is learned, as well as a cluster-specific 
mapping function that maps entities into the relation space. Thus, in CTransR, 
a relation will have multiple embeddings and multiple mapping functions so
as to capture the different subtypes among a same relation. The translation
among entities is same as that in TransR in the relation space. 
%

Existing work also tackles the mapping functions to embed entities and relations
within the translation based learning framework. 
In TransD~\citep{TransD}, a dynamic mapping matrix is calculated from the embeddings of
entities and relations to map entities from entity space to relation space. In the
relation space, a translation as in TransE is used to model the relations among
entities. 
TranSparse~\citep{TranSparse} uses relation-specific
sparse mapping matrices to map entities into the relation space, similarly as in TransR
and TransD. The sparsity of the mapping matrices varies according to the complexity of the relations.
Thus, complexities among the relations can be captured via mapping matrix sparsity, 
and meanwhile, overfitting risks are decreased. 
%
%
DistMult~\citep{DistMult} represents relations using diagonal matrices (e.g., $W_{r}$)
and use a bilinear scoring function to measure the distance from \head through \relation
to \tail (e.g., $\head W_{r} \tail^\intercal$).
%
%
In ComplEx~\citep{ComplEx}, the distance from \head through \relation to \tail is also measured
by dot product. However, instead of using a single representation for the entities and relations, 
ComplEx learns multiple representations for each entity and relation in different domains.
The distances from \head through \relation to \tail in the different domains 
are then combined (multiplied) as the final distance. 
HolE~\citep{HolE} can be considered as a special case of ComplEx, which models relations between 
entities using compositional vector representations based on circular correlations.
In ConvE~\citep{ConvE}, a 2D CNN architecture is used to convert $\e_{\scriptsize{\head}}$ and $\e_{\scriptsize{\relation}}$
to an estimated $\tilde{\e}_{\scriptsize{\tail}}$. The similarity between $\tilde{\e_{\scriptsize{\tail}}}$ and
the true tail embedding $\e_{\scriptsize{\tail}}$, via dot product, is used to calculate the probability score
of $(\head, \relation, \tail)$. Different from these methods mainly based on simple linear translations or 
correlations between (transformed) embeddings, our proposed methods learn a powerful nonlinear 
scoring metric of the $(\head, \relation, \tail)$ triplets, and directly model complex interactions between 
relation and entity embeddings via a CNN.

\begin{figure*}[!htbp]
  \centering
  \scalebox{0.85}[0.85]{
\begin{tikzpicture}[node distance = 1.2cm, thick, nodes = {align = center},
    >=latex]

\node [minimum width = {width("Magnetometer")+2pt}, fill = blue!50] (head)
    {head $\e_{\scriptsize{\head}}$};
\node [minimum width = {width("Magnetometer")+2pt}, fill = yellow!50, below = 0pt of head] (relation)
                           {relation $\e_{\scriptsize{\relation}}$};
\node [minimum width = {width("Magnetometer")+2pt}, fill = blue!50,   below = 0pt of relation] (tail)
    {tail $\e_{\scriptsize{\tail}}$};
\node [above=0pt of head] (M) {\textcolor{blue!70}{$m_{\scriptsize(\head, \relation, \tail)} \in \mathbb{R}^k$}};
\node [left=0pt of relation] (chain1) {primary chain}; 

%
\node [minimum width = {width("a")+2pt}, fill = red!30, thin,  right = of relation] (fmap1) {\\};
\node [minimum width = {width("a")+2pt}, fill = red!40, thin,  right = 0pt of fmap1] (fmap2)    {\\}; 
\node [minimum width = {width("a")+2pt}, fill = red!20, thin,  right = 0pt of fmap2] (fmap3)    {\\}; 
\node [above = 0pt of fmap2] (map)    {\textcolor{red!60}{$n_k$ feature}\\\textcolor{red!60}{maps in $\mathbb{R}^{k/3}$}};

\path(relation)edge[->] node[below, yshift=-20pt]{CNN\\($n_k$ kernels)} (fmap1);

\node [minimum width = {width("Magnetom")+2pt}, fill = green!30, right = of fmap3] (fflata) {$\fflata \in \mathbb{R}^{n_k\times k/3}$};
\node [above = 0pt of fflata] (ffltxt) {\textcolor{green!60}{flattened}\\\textcolor{green!60}{feature map}};
\path(fmap3)edge[->] node[below, yshift=-25pt]{flatten} (fflata);

\node [minimum width = {width("Magnetom")+2pt}, fill = orange!20, right = of fflata] (triplet) {$\ffcla$};
\node [above =0pt of triplet] (trip) {\textcolor{orange!60}{triplet}\\\textcolor{orange!60}{representation}};

\path(fflata) edge[->] node[below, yshift = -15pt]{fully-\\connected\\layers} (triplet);

\node [minimum size = 20pt, draw, thin, fill=violet!50, right = of triplet, shape = circle] (scorea) {\scorea};
\path(triplet)edge[->] node[below, yshift=-20pt]{logistic\\regression}(scorea);
%

\node [minimum width = {width("Magnetometer")+2pt}, fill = blue!20, below = 25pt of tail] (head_drop)
    {sparsified $\e_{\scriptsize{\head}}^s$};
\node [minimum width = {width("Magnetometer")+2pt}, fill = yellow!30, below = 0pt of head_drop] (relation_drop)
    {sparsified $\e_{\scriptsize{\relation}}^s$};
\node [minimum width = {width("Magnetometer")+2pt}, fill = blue!20, below = 0pt of relation_drop] (tail_drop)
    {sparsified $\e_{\scriptsize{\tail}}^s$};
\node [below=0pt of tail_drop] (Ms) {\textcolor{blue!50}{$m_{\scriptsize(\head, \relation, \tail)}^s \in \mathbb{R}^k$}};
\node [left=0pt of relation_drop] (chain1) {secondary chain}; 

%
\node [minimum width = {width("a")+2pt}, fill = purple!30, thin,  right = of relation_drop] (fmap1_drop) {\\};
\node [minimum width = {width("a")+2pt}, fill = purple!40, thin,  right = 0pt of fmap1_drop] (fmap2_drop)    {\\}; 
\node [minimum width = {width("a")+2pt}, fill = purple!20, thin,  right = 0pt of fmap2_drop] (fmap3_drop)    {\\}; 
\node [below = 0pt of fmap2_drop] (map_drop)    {\textcolor{purple!60}{$n_k$ feature}\\\textcolor{purple!60}{maps in $\mathbb{R}^{k/3}$}};

\path(relation_drop)edge[->] node[below]{} (fmap1_drop);
\path(relation_drop)edge[->] node[below, yshift=-5pt]{} (fmap1_drop);

\node [minimum width = {width("Magnetom")+2pt}, fill = cyan!30, right = of fmap3_drop] (fflata_drop) {$\fflatb \in \mathbb{R}^{n_k\times k/3}$};
\node [below = 0pt of fflata_drop] (ffltxt_drop) {\textcolor{cyan!60}{flattened}\\\textcolor{cyan!60}{feature map}};
\path(fmap3_drop)edge[->] node[below]{} (fflata_drop);

\node [minimum width = {width("Magnetom")+2pt}, fill = brown!20, right = of fflata_drop] (triplet_drop) {$\ffclb$};
\node [below =0pt of triplet_drop] (trip_drop) {\textcolor{brown!60}{triplet}\\\textcolor{brown!60}{representation}};

\path(fflata_drop) edge[->] (triplet_drop);

\node [minimum size = 20pt, draw, thin, fill=violet!30, right = of triplet_drop, shape = circle] (scoreb) {\scoreb};
\path(triplet_drop)edge[->] (scoreb);
%

\path(tail)edge[->] node[left]{drop out}(head_drop);

\node [below=28pt of fflata, rotate=90, yshift=-10pt] (drop_out){drop out}; 
\path(drop_out)edge[->] (fflata);
\path(drop_out)edge[->] (fflata_drop);

\node [below=30pt of triplet, rotate=90] (drop_out2){drop out}; 
\path(drop_out2)edge[->] (triplet);

\end{tikzpicture}
  }
  \caption{\method Model Architecture}
  \label{fig:method}
  \vspace{-10pt}
\end{figure*}
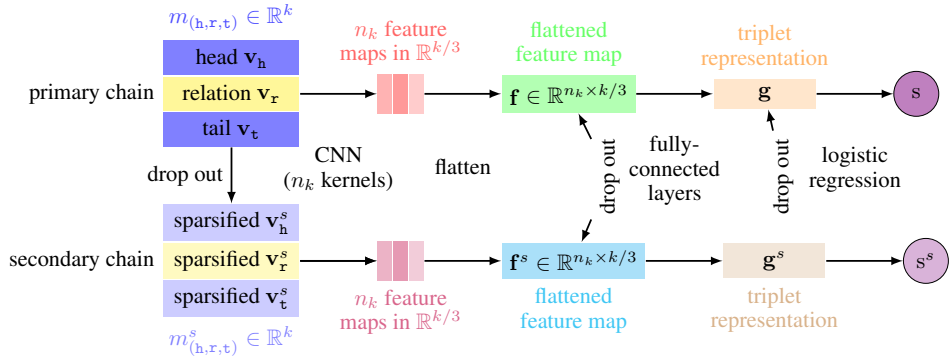

\subsection{{Knowledge Graph Learning with Additional Information}}
\label{sec:related:info}

Existing work also incorporates additional information (e.g., textual information of
entities, paths in knowledge graphs) to improve performance on knowledge graph learning. 
DKRL~\citep{DKRL} incorporates entity
descriptions in words, and applies a 2-layer CNN to map word descriptions to 
entity embeddings, in addition to learning another set of entity and relation embeddings
from the $(\head, \relation, \tail)$ triplets (i.e., knowledge graph structure).
DKRL leverages the translation idea from TransE, 
and enforces small distances between $\e_{\scriptsize{\head}} + \e_{\scriptsize{\relation}}$ and $\e_{\scriptsize{\tail}}$ for all
positive triplets, using
the embeddings learned from both word descriptions and knowledge graph structures. 
With the entity descriptions, DKRL is able to calculate embeddings for entities that are not
in its training data, and thus handle the zero-shot problem~\citep{DKRL}. 
%
%
CBOW~\citep{DKRL} is a variation of DKRL, in which the descriptions
are represented using the bag-of-word representation by 
their top-20 words weighted by TF-IDF. 
R-GCN~\citep{R-GCN} incorporates the path information of knowledge graphs. 
R-GCN learns the embedding for each entity in the knowledge graph via graph convolution~\citep{GCN},
and uses a similar idea as in DistMult to measure the relation between head and tail entities
(i.e., dot product between head embedding, diagonal matrix of relation and tail embedding). 
%
SSP~\citep{SSP} learns a topic model of the textual descriptions of entities. Unlike these previous approaches, our 
methods model entity descriptions with CNNs over word embeddings enhanced with a structural attention mechanism and 
perform knowledge graph completion in our proposed CNN based triplet scoring framework.
%

\section{CNN-based Dual-Chain Model (\method)}
\label{sec:method}
%

%

We develop a novel \underline{C}NN-based \underline{D}ual-\underline{C}hain model,
denoted as \method, for entity and relation prediction in knowledge graph. 
\method has two chains of learning processes: a primary chain and a secondary chain. 
In the primary chain of learning processes, each entity $\entity \in \eSet$ and each relation $\relation \in \rSet$
will be embedded into a low-dimensional vector $\e_{\entity} \in \mathbb{R}^k$
and $\e_{\scriptsize{\relation}} \in \mathbb{R}^k$, respectively, in a same latent space, where
$k$ is the embedding dimension. A CNN will be applied on the entity and relation embeddings
so as to capture their interactions, which will be used to predict (\head, \relation, \tail) triplets. 
We will learn such embeddings via supervised learning in \method such that
true (positive) $(\head, \relation, \tail)$ triplets will be scored high
and false (negative) triplets will be scored low.
A secondary chain of learning processes is designed to relax the
primary learning process so as to avoid overfitting, and meanwhile to strengthen
the major signals that will be learned from the primary chain in order to enforce
robust and consistent predictions. 
Figure~\ref{fig:method} presents the architecture of \method. 
%
%

\subsection{\method Model Architecture}
\label{sec:method:cnn}


%

%
\subsubsection{Dual Embedding Representations}
\label{sec:method:cnn:der}

In \method, given a triplet $(\head, \relation, \tail)$, we stack their embedding
vectors $\e_{\scriptsize{\head}}$, $\e_{\scriptsize{\relation}}$ and $\e_{\scriptsize{\tail}}$ into a matrix
$\m_{\scriptsize{(\head, \relation, \tail)}} \in \mathbb{R}^{3 \times k}$,
that is, $\m_{\scriptsize{(\head, \relation, \tail)}} = [\e_{\scriptsize{\head}}; \e_{\scriptsize{\relation}}; \e_{\scriptsize{\tail}}]$.
Meanwhile, we also use a sparsified version of $\m_{\scriptsize{(\head, \relation, \tail)}}$, 
denoted as $\m_{\scriptsize{(\head, \relation, \tail)}}^{s}$, by randomly dropping out
20\% of values in $\m_{\scriptsize{(\head, \relation, \tail)}}$. 
Both of $\m_{\scriptsize{(\head, \relation, \tail)}}$ and $\m_{\scriptsize{(\head, \relation, \tail)}}^{s}$
will go through exactly same network architectures with same parameters, and will be learned with shared parameters
in almost identical ways (except drop-out operations as indicated in Figure~\ref{fig:method}).
Therefore, we only discuss
the learning for $\m_{\scriptsize{(\head, \relation, \tail)}}$ in the following section.
The process with $\m_{\scriptsize{(\head, \relation, \tail)}}$/$\m_{\scriptsize{(\head, \relation, \tail)}}^s$
as input is referred to as the primary/secondary chain, respectively.

The use of the sparsified embeddings in addition to the original embeddings is inspired by
ComplEx~\citep{ComplEx}, 
which learns multiple embeddings for each entity to capture different aspects of the entity information. 
In our experiments, we observed that
learning multiple embeddings to represent each entity 
(i.e, replacing $\m_{\scriptsize{(\head, \relation, \tail)}}^s$ 
by another embedding that is also learned from the $(\head, \relation, \tail)$ triplet) will induce overfitting.
Instead, we use the sparsified version $\m_{\scriptsize{\head, \relation, \tail}}^s$ as a second
representation for $(\head, \relation, \tail)$.
Thus, the secondary chain is expected to still capture the major signals learned in 
$\m_{\scriptsize{\head, \relation, \tail}}$ and strengthen such signals as
$\m_{\scriptsize{\head, \relation, \tail}}^s$ goes through almost same learning process
as $\m_{\scriptsize{\head, \relation, \tail}}$ with same model parameters.
Meanwhile, the learning for the model parameters will be regularized by the drop-out
operations along the chains, thus the risk of overfitting is reduced.
The other drop-outs as in Figure~\ref{fig:method} will be discussed in
Section~\ref{sec:method:cnn:cfm}. 

\subsubsection{Convolution and Feature Mapping}
\label{sec:method:cnn:cfm}

\begin{figure}[!htbp]
  \vspace{-10pt}
  \centering
\begin{tikzpicture}[node distance = 1.2cm, thick, nodes = {align = center},
    >=latex]

  \node [minimum height = {height("x")+1em}, minimum width = {width("xxxxxxxxxxxxxxxxxxxxxxxxx")+2pt}, fill = blue!40] (head) {};
  \node [left=0pt of head] (headtxt) {$\e_{\scriptsize{\head}}$}; 
  \node [minimum height = {height("x")+1em}, minimum width = {width("xxxxxxxxxxxxxxxxxxxxxxxxx")+2pt}, fill = yellow!40, below = -1pt of head] (relation) {};
  \node [left=0pt of relation] (relationtxt) {$\e_{\scriptsize{\relation}}$}; 
  \node [minimum height = {height("x")+1em}, minimum width = {width("xxxxxxxxxxxxxxxxxxxxxxxxx")+2pt}, fill = blue!40,   below = -1pt of relation] (tail) {}; 
  \node [left=0pt of tail] (tailtxt) {$\e_{\scriptsize{\tail}}$};

  \node [right=-80pt of relation, minimum height = {height("x")+height("x")+height("x")+3em}, minimum width = {height("x")+height("x")+height("x")+3em}, fill = magenta!40, fill opacity=.7] (kernel) {};

  \node [right=-80pt of relation, thin, draw, minimum height = {height("x")+1em}, minimum width = {height("x")+1em}, fill = none, fill opacity=1] (c21) {};
  \node [right=-66pt of relation, thin, draw, minimum height = {height("x")+1em}, minimum width = {height("x")+1em}, fill = none, fill opacity=1] (c22) {};
  \node [right=-52pt of relation, thin, draw, minimum height = {height("x")+1em}, minimum width = {height("x")+1em}, fill = none, fill opacity=1] (c23) {};

  \node [above=-0.5pt of c21, thin, draw, minimum height = {height("x")+1em}, minimum width = {height("x")+1em}, fill = none, fill opacity=1] (c11) {};
  \node [above=-0.5pt of c22, thin, draw, minimum height = {height("x")+1em}, minimum width = {height("x")+1em}, fill = none, fill opacity=1] (c12) {};
  \node [above=-0.5pt of c23, thin, draw, minimum height = {height("x")+1em}, minimum width = {height("x")+1em}, fill = none, fill opacity=1] (c13) {};

  \node [below=-0.5pt of c21, thin, draw, minimum height = {height("x")+0.95em}, minimum width = {height("x")+1em}, fill = none, fill opacity=1] (c31) {};
  \node [below=-0.5pt of c22, thin, draw, minimum height = {height("x")+0.95em}, minimum width = {height("x")+1em}, fill = none, fill opacity=1] (c32) {};
  \node [below=-0.5pt of c23, thin, draw, minimum height = {height("x")+0.95em}, minimum width = {height("x")+1em}, fill = none, fill opacity=1] (c33) {};

  \node [above=5pt of c11] (text) {$3\times 3$ kernel};
  \node [right=0pt of text] (p1) {};
  \node [right=25pt of p1] (p2) {};
  \path (p1)edge[->] (p2); 
  
\end{tikzpicture}
  \caption{\method Convolution over Triplet Matrix}
  \label{fig:kernel}
  \vspace{-5pt}
\end{figure}
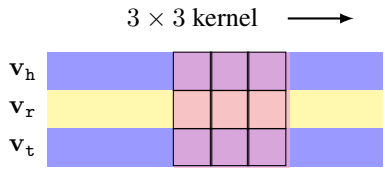

%
We first use $n_c$ kernels of size 3$\times$3 to conduct convolution over $\m_{\scriptsize{(\head, \relation, \tail)}}$
(and same for $\m_{\scriptsize{(\head, \relation, \tail)}}^s$). 
The convolution will go through the rows of $\m_{\scriptsize{(\head, \relation, \tail)}}$ 
as indicated in Figure~\ref{fig:kernel} 
%
%
Therefore, it will capture the interactions between \head, \relation and \tail since the
kernels have same height as $\m_{\scriptsize{(\head, \relation, \tail)}}$ and thus
cover and integrate information from \head, \relation and \tail at the same time. 
It is expected that the interaction patterns among positive triplets are different from
those among negative triplets, and thus lead to different label predictions. 
Out of the convolution, we will produce $n_k$ feature maps of size $k/3$$\times$1. 
We apply ReLU activation function, defined as $\text{ReLU}(x) = \max(0,x)$, on each of
feature maps to generate non-negative feature maps.
The ReLU activation is used here to only keep the non-negative interactions among the
triplets that will contribute to their labels, and meanwhile introduces non-linearality
in the feature mapping.
After the convolution operation, we flatten the feature maps (i.e., the output of the
convolution operation) into a feature vector $\fflata$ of size $1\times (n_kk/3)$ with 20\%
of the values randomly dropped out. 
We use a fully-connected layer (i.e., a projection matrix) with ReLU activation function
and map \fflata into another low-dimension feature $\ffcla$ as in Equation~\ref{eq:fcl}, 
\begin{equation}
  \label{eq:fcl}
  \ffcla = f(\fflata \Wfcl + \bfcl),
\end{equation}
where \Wfcl and \bfcl are the projection matrix and the bias, $f()$ is the
ReLU function.
After the mapping, \ffcla will have 20\% of its values randomly dropped out. 
The embedding $\m_{\scriptsize{(\head, \relation, \tail)}}^s$ will go through exactly the same operations
with same parameters, except that drop-out is only applied 
on the flattened feature maps.
The resulted feature from the fully-connected layer is denoted as $\ffclb$. 
Please note that the two drop-outs in the primary chain are to prevent
overfitting along the primary chain. In the secondary chain, since the input
triplets are already sparsified, we only applied drop-out on the flattened
feature maps, not on the triplet representations. Our experiments showed that
without any drop-outs on the secondary chain, the performance is suboptimal
compared to one drop-out on the secondary chain.

\subsubsection{Logistic Regression}
\label{sec:method:cnn:lr}

We consider the feature $\ffcla$ out of the fully-connected layer
as the feature representation of the triplet.
In \method, a positive triplet should be assigned a high score and
a negative triplet should have a low score. 
We use logistic regression to calculate the scores for the triplets based on $\ffcla$. 
The scoring function on $\ffcla$ is as in Equation~\ref{eqn:score}, 
\begin{equation}
  \label{eqn:score}
  \scorea(\head, \relation, \tail) = \sigma(\ffcla \Wlogit + \blogit)
  =\sigma(f(\fflata \Wfcl + \bfcl)\Wlogit + \blogit) , 
\end{equation}
where $\sigma$ is the sigmoid function,
$\Wlogit$ and $\blogit$ are the linear mapping weights and bias.  
Similarly, for $\m_{\scriptsize{(\head, \relation, \tail)}}^s$, the score out of logistic
regression is denoted as \scoreb. 
Both \scorea and \scoreb will be optimized as described in Section~\ref{sec:method:opt}
in order to learn the model.

\begin{figure*}[!h]
    \centering
    \begin{minipage}[b]{0.4\textwidth}
      \centering
      \resizebox {\columnwidth} {!} {
\begin{tikzpicture}[node distance = 1.2cm, thick, nodes = {align = center},
    >=latex]

\node [minimum width = {width("Magnetometer")+2pt}, fill = purple!40] (w1)    {$w_1$};
\node [minimum width = {width("Magnetometer")+2pt}, fill = purple!20, below = 0pt of w1] (w2) {$w_2$}; 
\node [minimum width = {width("Magnetometer")+2pt}, fill = purple!40, below = 0pt of w2] (w3) {$w_3$};
\node [minimum width = {width("Magnetometer")+2pt}, fill = purple!50, below = 0pt of w3] (w4) {$w_4$};
\node [minimum width = {width("Magnetometer")+2pt}, fill = purple!30, below = 0pt of w4] (w5) {$w_5$};

\node [above=0pt of w1] (M) {\textcolor{purple!70}{word embeddings ($D_{\scriptsize{\entity}}$)}}; 

%
\node [minimum width = {width("Magnetometer")+2pt}, fill = red!30, thin,  right = of w3] (fmap2) {$L_{\scriptsize{\entity}}$};
\node [minimum width = {width("Magnetometer")+2pt}, fill = blue!40, thin,  above = 0pt of fmap2] (fmap1)    {\textcolor{blue!40}{$v_1$}}; 
\node [minimum width = {width("Magnetometer")+2pt}, fill = green!20, thin,  below = 0pt of fmap2] (fmap3)    {\textcolor{green!20}{$v_3$}};

\node [above = 0pt of fmap1] (map)    {embedding matrix}; 
        
\path(w1.east) edge[red!30, ->] ($(fmap2.south west)!0.75!(fmap2.north west)$);
\path(w3.east) edge[red!30, ->] (fmap2.west);
\path(w5.east) edge[red!30, ->] ($(fmap2.south west)!0.25!(fmap2.north west)$);
\path(w1.east) edge[blue!40, ->] ($(fmap1.south west)!0.75!(fmap1.north west)$);
\path(w2.east) edge[blue!40, ->] (fmap1.west);
\path(w5.east) edge[blue!40, ->] ($(fmap1.south west)!0.25!(fmap1.north west)$);
\path(w1.east) edge[green!40, ->] ($(fmap3.south west)!0.75!(fmap3.north west)$);
\path(w4.east) edge[green!40, ->] (fmap3.west);
\path(w5.east) edge[green!40, ->] node[below, yshift=-10pt]{\textcolor{black}{structure}\\\textcolor{black}{attention ($A_{\entity}$)}} ($(fmap3.south west)!0.25!(fmap3.north west)$);

\node [minimum width = {width("x")+3pt}, fill = brown!30, right = of fmap2] (fflata) {\\ \\ \\ \\};
\node [above = 0pt of fflata] (ffltxt) {\textcolor{brown!70}{feature}\\\textcolor{brown!70}{map}}; 

\path(fmap2)edge[->] node[below, yshift=-10pt]{conv\\\&max\\pooling} (fflata);



\node [minimum width = {width("Magnetom")+2pt}, minimum height = 0.5em, fill = teal!20, thin,  right = of fflata] (embedding) {textual \etxt};
\node [above=0pt of embedding] (embtxt) {\textcolor{teal!80}{entity embedding}\\\textcolor{teal!80}{from descriptions}}; 
\path(fflata) edge[->] node[below, yshift = -10pt]{conv\\\&mean\\pooling} (embedding);

\end{tikzpicture}
}
      \caption{\mbox{Entity Embeddings from Descriptions in \methodD}}
      \label{fig:mapping_txt}
    \end{minipage}\hfill
    \begin{minipage}[b]{0.56\textwidth}
      \centering
      \resizebox {\columnwidth} {!} {
\begin{tikzpicture}[node distance = 1.2cm, thick, nodes = {align = center},
    >=latex]

\node [minimum width = {width("Magnetometer")+2pt}, fill = blue!50] (head)
    {head $\e_{\scriptsize{\head}}$};
\node [minimum width = {width("Magnetometer")+2pt}, fill = yellow!50, below = 0pt of head] (relation)
                           {relation $\e_{\scriptsize{\relation}}$};
\node [minimum width = {width("Magnetometer")+2pt}, fill = blue!50,   below = 0pt of relation] (tail)
    {tail $\e_{\scriptsize{\tail}}$};
\node [above=0pt of head] (M) {\textcolor{blue!70}{$m_{\scriptsize(\head, \relation, \tail)} \in \mathbb{R}^k$}};
\node [left=40pt of head, rotate=90, xshift=10pt, yshift=-20pt] (str) {structure\\information}; 

%
\node [minimum width = {width("a")+2pt}, fill = red!30, thin,  right = of relation] (fmap1) {\\};
\node [minimum width = {width("a")+2pt}, fill = red!40, thin,  right = 0pt of fmap1] (fmap2)    {\\}; 
\node [minimum width = {width("a")+2pt}, fill = red!20, thin,  right = 0pt of fmap2] (fmap3)    {\\}; 
\node [above = 0pt of fmap2] (map)    {\textcolor{red!60}{$n_k$ feature}\\\textcolor{red!60}{maps in $\mathbb{R}^{k/3}$}};

\path(relation)edge[->] node[below, yshift=-20pt]{CNN\\$n_k$ kernels} (fmap1);

\node [minimum width = {width("Magnetom")+2pt}, fill = green!30, right = of fmap3] (fflata) {$\fflatstr \in \mathbb{R}^{n_kk/3}$};
\node [above = 0pt of fflata] (ffltxt) {\textcolor{green!60}{flattened}\\\textcolor{green!60}{feature map}};
\path(fmap3)edge[->] node[below, yshift=-25pt]{flatten} (fflata);

\node [minimum width = {width("Magnetom")+2pt}, fill = orange!20, right = of fflata] (triplet) {$\ffclstr$};
\node [above =0pt of triplet] (trip) {\textcolor{orange!60}{triplet}\\\textcolor{orange!60}{representation}};

\path(fflata) edge[->] node[below, yshift = -15pt]{fully-\\connected\\layers} (triplet);

\node [minimum size = 20pt, draw, thin, fill=violet!50, right = of triplet, shape = circle] (scorestr) {\scorestr};
\path(triplet)edge[->] node[below, yshift=-20pt]{logistic\\regression}(scorestr);
%

\node [minimum width = {width("Magnetometer")+2pt}, fill = teal!20, below = 25pt of tail] (head_txt)
    {textual $\e_{\scriptsize{\head}}^d$};
\node [minimum width = {width("Magnetometer")+2pt}, fill = lime!30, below = 0pt of head_txt] (relation_txt)
    {textual $\e_{\scriptsize{\relation}}^d$};
\node [minimum width = {width("Magnetometer")+2pt}, fill = teal!20, below = 0pt of relation_txt] (tail_txt)
    {textual $\e_{\scriptsize{\tail}}^d$};
\node [below=0pt of tail_txt] (Ms) {\textcolor{teal!50}{$m_{\scriptsize(\head, \relation, \tail)}^t \in \mathbb{R}^k$}};
\node [left=40pt of head_txt, rotate=90, xshift=10pt, yshift=-20pt] (str) {entity\\descriptions}; 
%
\node [minimum width = {width("a")+2pt}, fill = purple!30, thin,  right = of relation_txt] (fmap1_txt) {\\};
\node [minimum width = {width("a")+2pt}, fill = purple!40, thin,  right = 0pt of fmap1_txt] (fmap2_txt)    {\\}; 
\node [minimum width = {width("a")+2pt}, fill = purple!20, thin,  right = 0pt of fmap2_txt] (fmap3_txt)    {\\}; 
\node [below = 0pt of fmap2_txt] (map_txt)    {\textcolor{purple!60}{$n_k$ feature}\\\textcolor{purple!60}{maps in $\mathbb{R}^{k/3}$}};

\path(relation_txt)edge[->] node[below]{} (fmap1_txt);
\path(relation_txt)edge[->] node[below, yshift=-5pt]{} (fmap1_txt);

\node [minimum width = {width("Magnetom")+2pt}, fill = cyan!30, right = of fmap3_txt] (fflata_txt) {$\fflattxt \in \mathbb{R}^{n_kk/3}$};
\node [below = 0pt of fflata_txt] (ffltxt_txt) {\textcolor{cyan!60}{flattened}\\\textcolor{cyan!60}{feature map}};
\path(fmap3_txt)edge[->] node[below]{} (fflata_txt);

\node [minimum width = {width("Magnetom")+2pt}, fill = brown!20, right = of fflata_txt] (triplet_txt) {$\ffcltxt$};
\node [below =0pt of triplet_txt] (trip_txt) {\textcolor{brown!60}{triplet}\\\textcolor{brown!60}{representation}};

\path(fflata_txt) edge[->] (triplet_txt);

\node [minimum size = 20pt, draw, thin, fill=violet!30, right = of triplet_txt, shape = circle] (scoretxt) {\scoretxt};
\path(triplet_txt)edge[->] (scoretxt);

\node [below=30pt of triplet, rotate=90] (drop_out2){drop out}; 
\path(drop_out2)edge[->] (triplet);
\path(drop_out2)edge[->] (triplet_txt);

\node [below=15pt of scorestr, xshift=15pt, minimum size = 20pt, draw, thin, fill=violet!70, shape = circle] (score) {\score};
\path (scorestr) edge[->] (score);
\path (scoretxt) edge[->] (score); 

\end{tikzpicture}
}
      \vspace{-20pt}
      \caption{\methodD Model Architecture}
      \label{fig:architecture_txt}
    \end{minipage}
    \vspace{-15pt}
\end{figure*}

\subsection{The \method Optimization Problem}
\label{sec:method:opt}
%
We use entropy loss as the objective in order to learn the \method model.
As mentioned in Section~\ref{sec:method:cnn:lr} and shown in Figure~\ref{fig:method}, 
\method will produce two scores \scorea and \scoreb for each triplet.
During training, we optimize these two scores jointly.
Thus, the objective function for \method is as in Equation~\ref{eqn:entropy},
\begin{equation}
  \label{eqn:entropy}
  \begin{aligned}
    \min_{\boldsymbol{\Theta}}
    \sum_{\substack{\scriptsize{(\head, \relation, \tail)\in \tSet}}}\log(\scorea(\head, \relation, \tail)) +
    \sum_{\substack{\scriptsize{(\head',\relation', \tail')\in \tSet'}}} \log(1-\scorea(\head', \relation', \tail')) + \\
    \sum_{\substack{\scriptsize{(\head, \relation, \tail)\in \tSet}}}\log(\scoreb(\head, \relation, \tail)) +
    \sum_{\substack{\scriptsize{(\head',\relation', \tail')\in \tSet'}}} \log(1-\scoreb(\head', \relation', \tail')),
  \end{aligned}
\end{equation}
where $\boldsymbol{\Theta} = \{ \m_{\scriptsize{(\head, \relation, \tail)}}, 
 \bfcl, \Wfcl, \blogit, \Wlogit, \mathcal{K}\}$ is the set of parameters in \method
 and $\mathcal{K}$ is the set of kernels used for convolution,
\tSet is the set of positive triplets and \tSet' is manually corrupted negative set
as will be described in Section~\ref{sec:method:opt:learn}, and \scorea and \scoreb are defined as in
Equation~\ref{eqn:score}.
That is, \method makes predictions from two input embeddings, and optimizes both of the two
predicted scores to be close to the ground truth. 
The score \scorea is used to predict the relations among triplets during testing. 

The reason why only \scorea, instead of both \scorea and \scoreb, is used for prediction is that,
when testing, the drop-out rate will be 0. 
As a result, the secondary chain will also start from 
the same entity and relation embeddings as in the primary chain.
Since the architectures of the two chains are identical, 
\scorea and \scoreb calculated from the two chains will be identical. 
For simplicity, we use \scorea for prediction in \method. 
%
%
%
However, during model learning, both \scorea and \scoreb are used in the objective
function so \scoreb can enhance \scorea for better optimization. We also observed
from our experiments that the performance without \scoreb in the objective function
is worse than that with both the scores. 
The reason why we use entropy loss instead of pairwise ranking loss is that
we observed tie problems from pairwise ranking loss in our experiments, that is,
multiple triplets may have exactly same scores and thus be ranked equally, which
causes overestimate using ranking based metric (e.g., hits at 10 as will be
discussed in Section~\ref{sec:exp:eval}). 
%

\subsection{\method Learning Algorithm and Negative Triplet Generation}
\label{sec:method:opt:learn}

%
%
%
We initialize the embeddings and the model parameters using
uniform distributions over $[-\sqrt{\frac{6}{n_{\text{in}} + n_{\text{out}}}}, \sqrt{\frac{6}{n_{\text{in}} + n_{\text{out}}}}]$~{\citep{glorot2010understanding}},
where $n_{\text{in}}$ is the in dimension, $n_{\text{out}}$ is the out dimension
of the embeddings or parameters
(e.g., for a matrix $X \in \mathbb{R}^{m \times n}$, 
$m$ is the in dimension, $n$ is the out dimension). 
At each epoch, multiple batches are created and used as training data, and 
mini-batch stochastic gradient descent method is used to update all the parameters.
The training batches are generated as follows.  
We first randomly choose $n_b$ positive triplets from positive training set \tSet.
For every positive triplet $(\head, \relation, \tail)$, we generate a corresponding
negative triplet by randomly replacing either \head or \tail (but not both at same
time) by $\head'$ or $\tail'$, where $(\head', \relation, \tail) \notin \tSet$ and
$(\head, \relation, \tail') \notin \tSet$.
%
The negative triplets are sampled for each batch of positive triplets. 
We use the distribution described in 
Wang \etal~\citep{TransH}
to decide to replace head or tail.
Therefore, there will be $n_b$ pairs of positive and negative triplets in a batch. 
We use the batches of $n_b$ pairs of triplets for \method model learning. 
%
%
%

\section{\methodD with Entity Descriptions}
\label{sec:methodd}
%
%
When entity descriptions are available, they may provide additional, useful information
that can help relation prediction over entities. 
%
%
Thus, we extend \method to incorporate available entity descriptions, 
and denote the new model as \methodD.
Given the entity descriptions in the form of word vectors,
\methodD learns a mapping function from entity descriptions to entity embeddings. 
Figure~\ref{fig:mapping_txt} presents the scheme of such mapping. 
%
%
Structure attention mechanism~\citep{StructrueAtt} and two-layers CNN~\citep{DKRL} are applied to learn the entity
embeddings from descriptions, denoted as \etxt.  
%

\subsection{Attention and Convolution over Word Embeddings}
\label{sec:methodd:txt:str_att}

We use the structure attention mechanism~{\citep{StructrueAtt}} to 
map the word embeddings of the descriptions of an entity, denoted as $D_{\scriptsize{\entity}}$,
to a more expressive embedding matrix denoted as $L_{\scriptsize{\entity}}$.
The attention weights are calculated from $D_{\scriptsize{\entity}}$ as follows,   
\begin{equation}
  \label{eqn:str:att}
  A_{\scriptsize{\entity}} = \text{softmax}(V\tanh(UD_{\scriptsize{\entity}}^\intercal)),
\end{equation}
where $U$ and $V$ are two weighting matrices shared across
  all the entities. 
$A_{\scriptsize{\entity}}$ is the attention weight matrix, which 
%
maps $D_{\scriptsize{\entity}}$ to $L_{\scriptsize{\entity}}$ as follows,
\begin{equation}
  \label{eqn:str:map}
  L_{\scriptsize{\entity}} = A_{\scriptsize{\entity}}D_{\scriptsize{\entity}}. 
\end{equation}
Intuitively, each row of $L_{\scriptsize{\entity}}$ represents one semantic aspect of the
entity \entity's description.

We use a two-layer CNN to map the $L_{\scriptsize{\entity}}$ to an entity embedding $\etxt$.
In the first layer, a convolution kernel is applied on $L_{\scriptsize{\entity}}$, followed by a max pooling.
The output from this layer goes through a second layer of convolution, followed by
a mean pooling. The output will be reshaped into $\e_{\scriptsize{\entity}}^t$ for the entity \entity. 

%

\subsection{Learning with Entity Descriptions}
\label{sec:method:txt:optimize}

Given the descriptions, we are able to produce two heterogeneous
embeddings for each entity.
One is from the structure information (i.e., relations among entities)
and the other is from the textual information (i.e., entity descriptions in words). 
We use the architecture shown in Figure~\ref{fig:architecture_txt} to calculate
two scores \scorestr and \scoretxt from these two embeddings, respectively, and
use the mean of these two scores as the output from the architecture. This
method is denoted as \methodD. 
That is, \methodD replaces the sparsified version of the entity embeddings in \method
(Figure~\ref{fig:method}) in the inputs by the entity description embeddings
(Figure~\ref{fig:mapping_txt}), 
and the two scores at the output in \method by the mean
of two scores.
Also, \methodD applies drop-outs over the triplet representations, 
with all the other architectures in \method remained same. 
The reason why the mean of \scorestr and \scoretxt is used as the output is that
we consider the embeddings from the entity descriptions 
and knowledge graph structure information both important in contributing to the final 
prediction, as they may encode different, complementary information about the entities. 

In \methodD, we denote the triplet representations (similar as 
\ffcla and \ffclb in \method) learned from structure information and textual information 
as \ffclstr and \ffcltxt, respectively.
We enforce the similarity between \ffclstr and \ffcltxt by introducing an $\ell_1$-norm penalty
in \methodD model training so that the structure information and the textual information can
be better fused in the model. 
%
We use $\ell_1$ norm here due to its empirical superior performance over other norms
(e.g., $\ell_2$ norm) in similar applications~{\citep{DKRL}}.
Therefore, the objective function for \methodD is as follows, 
\begin{eqnarray}
  \label{obj:txt_str}
  \begin{aligned}
    & \min_{\boldsymbol{\Theta}} && \sum_{\substack{\scriptsize{(\head, \relation, \tail)\in \tSet}}}\log(\score(\head, \relation, \tail))
    + \sum_{\substack{\scriptsize{(\head',\relation', \tail')\in \tSet'}}} \log(1-\score(\head', \relation', \tail')) \\
    &                            && + \sum_{\scriptsize{\tSet \cup \tSet'}}\lVert \ffcltxt - \ffclstr \rVert_{\ell_1}, \\
    & \text{s.t.,}               && \score(\head, \relation, \tail) = 0.5\times(\scoretxt(\head, \relation, \tail) 
                                   + \scorestr(\head, \relation, \tail)), \\
    &                            &&    \score(\head', \relation', \tail') = 0.5\times(\scoretxt(\head', \relation', \tail') 
    + \scorestr(\head', \relation', \tail')), \\
  \end{aligned}
\end{eqnarray}
where $\boldsymbol{\Theta} = \{U, V, D_{\scriptsize{\entity}}, \Theta_{\text{CNN}}, \m_{\scriptsize{(\head, \relation, \tail)}}, 
 \bfcl, \Wfcl, \blogit, \Wlogit, \mathcal{K}\}$, where 
 $\Theta_{\text{CNN}}$ is the set of parameters of the two-layer CNN including the kernels and bias,
 and $\mathcal{K}$ is the set of kernels as in \method. 
 The learning algorithm for \methodD is same to that for \method as in Section~\ref{sec:method:opt:learn}.

\section{Comparison between \method and Existing Methods}
\label{sec:comp}
 
\begin{figure*}[!h]
    \centering
    \begin{minipage}[b]{0.75\textwidth}
      \centering
      \includegraphics[width=\linewidth]{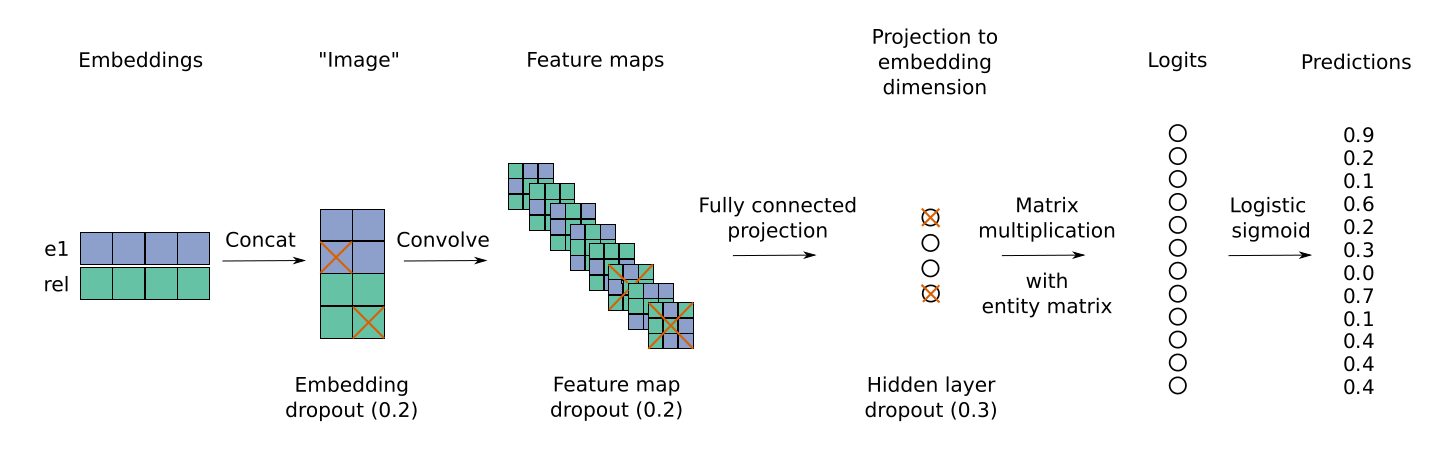}
      \vspace{-20pt}
      \caption{ConvE Model Architecture}
      \label{fig:ConvE}
    \end{minipage}\hfill
    \begin{minipage}[b]{0.25\textwidth}
      \centering
\begin{tikzpicture}[node distance = 0.3cm, thick, nodes = {align = center}, 
    >=latex]

  \node [thin, draw, minimum height = {height("x")/3+0.5em}, minimum width = {10 * (height("x")/3+0.5em)}, fill = blue!50, fill opacity=1] (h1) {};
  \node [thin, draw, minimum height = {height("x")/3+0.5em}, minimum width = {10 * (height("x")/3+0.5em)}, fill = blue!50, below = -1pt of h1] (h2) {};
  \node [thin, draw, minimum height = {height("x")/3+0.5em}, minimum width = {10 * (height("x")/3+0.5em)}, fill = blue!50, below = -1pt of h2] (h3) {};
  \node [thin, draw, minimum height = {height("x")/3+0.5em}, minimum width = {10 * (height("x")/3+0.5em)}, fill = blue!50, below = -1pt of h3] (h4) {};
  \node [thin, draw, minimum height = {height("x")/3+0.5em}, minimum width = {10 * (height("x")/3+0.5em)}, fill = blue!50, below = -1pt of h4] (h5) {};
  \node [left=0pt of h3] (headtxt) {$e1$};

  \node [thin, draw, minimum height = {height("x")/3+0.5em}, minimum width = {10 * (height("x")/3+0.5em)}, fill = green!50, below = -1pt of h5] (r1) {};
  \node [thin, draw, minimum height = {height("x")/3+0.5em}, minimum width = {10 * (height("x")/3+0.5em)}, fill = green!50, below = -1pt of r1] (r2) {};
  \node [thin, draw, minimum height = {height("x")/3+0.5em}, minimum width = {10 * (height("x")/3+0.5em)}, fill = green!50, below = -1pt of r2] (r3) {};
  \node [thin, draw, minimum height = {height("x")/3+0.5em}, minimum width = {10 * (height("x")/3+0.5em)}, fill = green!50, below = -1pt of r3] (r4) {};
  \node [thin, draw, minimum height = {height("x")/3+0.5em}, minimum width = {10 * (height("x")/3+0.5em)}, fill = green!50, below = -1pt of r4] (r5) {};
  \node [left=0pt of r3] (headtxt) {$rel$};

  \node [above left = -1pt and -(height("x")/3+0.5em)*1.5 of h1] (H1u) {};
  \node [below left = -1pt and -(height("x")/3+0.5em)*1.5 of r5] (H1d) {};
  \draw[thin, black] (H1u) -- (H1d);

  \node [above left = -1pt and -(height("x")/3+0.5em)*2.5 of h1] (H2u) {};
  \node [below left = -1pt and -(height("x")/3+0.5em)*2.5 of r5] (H2d) {};
  \draw[thin, black] (H2u) -- (H2d);

  \node [above left = -1pt and -(height("x")/3+0.5em)*3.5 of h1] (H3u) {};
  \node [below left = -1pt and -(height("x")/3+0.5em)*3.5 of r5] (H3d) {};
  \draw[thin, black] (H3u) -- (H3d);

  \node [above left = -1pt and -(height("x")/3+0.5em)*4.5 of h1] (H4u) {};
  \node [below left = -1pt and -(height("x")/3+0.5em)*4.5 of r5] (H4d) {};
  \draw[thin, black] (H4u) -- (H4d);

  \node [above left = -1pt and -(height("x")/3+0.5em)*5.5 of h1] (H5u) {};
  \node [below left = -1pt and -(height("x")/3+0.5em)*5.5 of r5] (H5d) {};
  \draw[thin, black] (H5u) -- (H5d);

  \node [above left = -1pt and -(height("x")/3+0.5em)*6.5 of h1] (H6u) {};
  \node [below left = -1pt and -(height("x")/3+0.5em)*6.5 of r5] (H6d) {};
  \draw[thin, black] (H6u) -- (H6d);

  \node [above left = -1pt and -(height("x")/3+0.5em)*7.5 of h1] (H7u) {};
  \node [below left = -1pt and -(height("x")/3+0.5em)*7.5 of r5] (H7d) {};
  \draw[thin, black] (H7u) -- (H7d);

  \node [above left = -1pt and -(height("x")/3+0.5em)*8.5 of h1] (H8u) {};
  \node [below left = -1pt and -(height("x")/3+0.5em)*8.5 of r5] (H8d) {};
  \draw[thin, black] (H8u) -- (H8d);

  \node [above left = -1pt and -(height("x")/3+0.5em)*9.5 of h1] (H9u) {};
  \node [below left = -1pt and -(height("x")/3+0.5em)*9.5 of r5] (H9d) {};
  \draw[thin, black] (H9u) -- (H9d); 
  

  \node [left=-(height("x")/3+0.5em)*6 of h5, minimum height = {height("x")/3+height("x")/3+height("x")/3+1.5em}, minimum width = {height("x")/3+height("x")/3+height("x")/3+1.5em}, 
  fill = magenta!40, fill opacity=.7] (kernel) {};

  \node [left=-(height("x")/3+0.5em)*4 of h5, thin, draw, minimum height = {height("x")/3+0.5em}, minimum width = {height("x")/3+0.5em}, fill = none, fill opacity=1] (c21) {};
  \node [left=-(height("x")/3+0.5em)*5 of h5, thin, draw, minimum height = {height("x")/3+0.5em}, minimum width = {height("x")/3+0.5em}, fill = none, fill opacity=1] (c22) {};
  \node [left=-(height("x")/3+0.5em)*6 of h5, thin, draw, minimum height = {height("x")/3+0.5em}, minimum width = {height("x")/3+0.5em}, fill = none, fill opacity=1] (c23) {};

  \node [above=-0.5pt of c21, thin, draw, minimum height = {height("x")/3+0.5em}, minimum width = {height("x")/3+0.5em}, fill = none, fill opacity=1] (c11) {};
  \node [above=-0.5pt of c22, thin, draw, minimum height = {height("x")/3+0.5em}, minimum width = {height("x")/3+0.5em}, fill = none, fill opacity=1] (c12) {};
  \node [above=-0.5pt of c23, thin, draw, minimum height = {height("x")/3+0.5em}, minimum width = {height("x")/3+0.5em}, fill = none, fill opacity=1] (c13) {};

  \node [below=-0.5pt of c21, thin, draw, minimum height = {height("x")/3+0.475em}, minimum width = {height("x")/3+0.5em}, fill = none, fill opacity=1] (c31) {};
  \node [below=-0.5pt of c22, thin, draw, minimum height = {height("x")/3+0.475em}, minimum width = {height("x")/3+0.5em}, fill = none, fill opacity=1] (c32) {};
  \node [below=-0.5pt of c23, thin, draw, minimum height = {height("x")/3+0.475em}, minimum width = {height("x")/3+0.5em}, fill = none, fill opacity=1] (c33) {};

  \node [below=1pt of r1] (text) {$3\times 3$ kernel};
  
\end{tikzpicture}
      \caption{ConvE Convolution over Head and Relation Embeddings}
      \label{fig:interactions}
    \end{minipage}
    \vspace{-25pt}
\end{figure*}

The model that is most similar to \method is ConvE~\citep{ConvE}, 
which employs 2D CNNs to learn embeddings for entities and relations.
However, there are several key differences between \method and ConvE.
The ConvE model architecture is shown in Figure~\ref{fig:ConvE} (extracted from the
ConvE paper~\citep{ConvE}).

%
In ConvE, the embedding of head (e.g., $e1$ in Figure~\ref{fig:ConvE})
and relation (e.g., $rel$ in Figure~\ref{fig:ConvE}) are stacked and
reshaped into a matrix (e.g., of size 10$\times$20, the matrix under ``Image'' in
Figure~\ref{fig:ConvE}), denoted as $e1$ and $rel$ respectively.
Then, a CNN along with a fully-connected layer is employed to project the head
and relation representations together into an embedding (e.g., the embedding under ``Projection
to embedding dimension'' in Figure~\ref{fig:ConvE}).
The probabilities of the projected
embedding belonging to different tail entities are calculated via the 
dot product between the projected embedding and the corresponding entity embeddings
and logits
(e.g., the probabilities under ``Predictions'' in Figure~\ref{fig:ConvE}). 
The key idea of ConvE is similar to TransE~\citep{TransE}, that is, to ``translate'' head and
relation to tail,
except that ConvE uses a CNN rather than a simple linear operation as in TransE
to model the translation.
In ConvE, multiple kernels are used in the CNN to capture the interactions between head and
relation. 
However, the kernels are typically very small (e.g., 3$\times$3 for 10$\times$20 reshaped
matrices).
Therefore, the kernels can only capture the interactions at the boundaries of the head
and relation embeddings as demonstrated in Figure~\ref{fig:interactions}.
However, in \method, the CNN can directly capture the interactions from the entirety of the
head, relation and tail embeddings as demonstrated in Figure~\ref{fig:kernel}. 

\section{Experimental Protocols}
\label{sec:exp}

\subsection{Datasets and Comparison Methods}
\label{sec:exp:data}

We use the following 8 public datasets in our experiments: 
Yago3-10~\citep{YAGO3}, WN18~\citep{TransE}, FB15k~\citep{TransE},
FB14k~\citep{DKRL}, FB15k-237~\citep{ConvE}, FB20k~\citep{DKRL},
DGI~\citep{DGIdb} and DDI~\citep{twosides}. 
\begin{itemize}[noitemsep,nolistsep,leftmargin=*]
\item \emph{YAGO3-10 dataset} contains relations among people and their attributes
  (e.g., (``Alex'', ``is a'', ``football player'')), and each relation has more than 10 triplets. 
%
\item \emph{WN18} is a subset of WordNet~\citep{wordnet}. 
Most triplets in WN18 represent the semantic relation between words.
\item \emph{FB15k dataset} is a subset of Freebase~\citep{bollacker2008freebase}.
Most triplets record facts about actors, awards, sports, sport teams and movies, and their relations. 
\item \emph{FB14k dataset} is a subset of FB15k, 
with entities that do not have descriptions removed.
%
\item \emph{FB15k-237 dataset} is another subset of FB15k,
where inverse relations are removed. 
%
\item \emph{FB20k dataset} is an extension of FB14k.
It contains 19,923 entities, among which 5,019 entities never appear in the training set.
\item \emph{DGI dataset} is a subset of drug-gene interaction dataset described in
Cotto \etal~{\citep{DGIdb}}
  with interaction types of frequency higher than 50.
  Common drug-gene interaction types include activation and inhibition, etc. 
\item \emph{DDI dataset} is a subset of drug-drug interaction dataset TWOSIDES~\citep{twosides} with
  interaction types of frequency between 500 and 1,100. Relations in DDI represent
  side effects that are induced by drug-drug interactions. 
  Common drug-drug interaction (i.e., side effect) types include food intolerance, breast cyst and bunion,
  etc.
\end{itemize}
The first 6
datasets are widely used benchmark datasets
in knowledge graph learning. 
%
The last 2 datasets are biological datasets for important biological problems.
It's noteworthy that 
FB15k and WN18 are reported to have test leakage issues~\citep{ConvE}.
Although FB15k and WN18 are widely used, they 
may not be good benchmarks for the evaluation purpose.
Table~\ref{tbl:data} presents the statistics of the datasets.
\vspace{-2pt}
\begin{table}[!ht]
  \caption{Dataset Statistics}
  \label{tbl:data}
  \centering
  \begin{threeparttable}
      \begin{tabular}{
	@{\hspace{12pt}}l@{\hspace{12pt}}
	@{\hspace{12pt}}r@{\hspace{12pt}}          
	@{\hspace{12pt}}r@{\hspace{12pt}}
        @{\hspace{12pt}}r@{\hspace{12pt}}
        @{\hspace{12pt}}r@{\hspace{12pt}}
        @{\hspace{12pt}}r@{\hspace{12pt}}
	}
        \toprule
        dataset  & \#entity & \#relation & \#train & \#valid & \#test \\
        \midrule
        YAGO3-10 & 123,182 & 37 & 1,079,040 & 5,000 & 5,000 \\
        WN18 & 40,943 & 18 & 141,442 & 5,000 & 5,000 \\
        FB15k & 14,951 & 1,345 & 483,142 & 50,000 & 59,071 \\
        FB14k & 14,904 & 1,341 & 472,860 & 48,991 & 57,809 \\
        FB15k-237 & 14,541 & 237 & 272,115 & 17,535 & 20,466 \\
        FB20k & 19,923 & 1,341 & 472,860 & 48,991 & 30,490 \\    
        DGI & 7,217 & 14 & 15,781 & 1,130 & 1,131 \\
        DDI & 538 & 25 & 20,951 & 2,618 & 2,617 \\
        \bottomrule
      \end{tabular}
      \begin{tablenotes}
        \setlength\labelsep{0pt}
	\begin{footnotesize}
	\item
          The columns corresponding to \#entity, \#relation, \#train, 
          \#valid and \#test have the number of entities, the number of
          relations, the number of training, validation and testing
          triplets, respectively. 
          \par
	\end{footnotesize}
      \end{tablenotes}
  \end{threeparttable}
\end{table}

\vspace{-1pt}

We compare \method with the following 
15 methods in total, where
the results of these methods on the datasets are available from literature:
TransE~\citep{TransE}, TransH~\citep{TransH},
TransR~\citep{TransR}, CTransR~\citep{TransR}, KG2E~\citep{KG2E}, TransD~\citep{TransD},
TranSparse~\citep{TranSparse}, HolE~\citep{HolE}, DistMult~\citep{DistMult}, ComplEx~\citep{ComplEx}, R-GCN~\citep{R-GCN}.
We did not compare with ConvE due to the data unfairness and reproducibility issues
as discussed later in Appendix~\ref{sec:comp:repro}. We include the corresponding results
in Table~\ref{tbl:compare_with_ConvE} in appendix.
%
We compare \methodD with 
DKRL~\citep{DKRL} and SSR~\citep{SSP}, which are able to use entity descriptions and have
results reported on FB14k. 

%

\subsection{Evaluation Metrics}
\label{sec:exp:eval}
%
%
%
We use two popular evaluation metrics: mean rank (\MRank)~{\citep{TransE}}
and hits at 10 (\HITten)~{\citep{TransE}},
to evaluate the performance of the methods, calculated as follows. 
%
%
%
%
For each positive triplet in the testing set, 
we generate a set of negative triplets by 
corrupting its head by each of the other entities in \eSet, but remove the negative
triplets if they are in training or validation set.
We predict the scores for each positive triplet and its corresponding negative triplets, 
and rank all the triplets using their predicted scores in descending order. 
\MRank is calculated as the average ranking position of all the positive testing triplets among
their corresponding negative triplets.
Low \MRank indicates that true positive triplets are ranked high, and thus
good performance. 
\HITten is calculated as the fraction of the positive testing triplets over all positive
testing triplets that are ranked into top 10 among their corresponding negative testing
triplets.
High \HITten indicates that many true positive triplets are ranked among top 10 of their negative
triplets, and thus good performance.

\section{Experimental Results}
\label{sec:results}

\subsection{Entity Prediction}
\label{sec:results:entity}

For all the testing triplets (\head, \relation, \tail), we conducted two sets of experiments:
1) prediction of \tail given \head and \relation, and 2) prediction of \head given \tail and \relation.
We evaluated the performance of each experiment using \MRank and \HITten, and present
the respective average results over the two sets of the experiments in Table~\ref{tbl:results_wn18_fb15k},
\ref{tbl:results_YAGO3_FB15K-237} and \ref{tbl:results_dgi_ddi}.

\begin{table}[!h]
  \caption{Entity Prediction on WN18 and FB15k}
  \label{tbl:results_wn18_fb15k}
  \centering
  \begin{threeparttable}
    \begin{tabular}{	 
    	  @{\hspace{8pt}}l@{\hspace{18pt}}
	  @{\hspace{18pt}}r@{\hspace{18pt}}
          @{\hspace{18pt}}r@{\hspace{18pt}} 
	  @{\hspace{18pt}}c@{\hspace{18pt}}
          @{\hspace{18pt}}r@{\hspace{18pt}}
	  @{\hspace{18pt}}r@{\hspace{8pt}}
	}	
	\toprule
        \multirow{2}{*}{method} & \multicolumn{2}{c}{WN18} & & \multicolumn{2}{c}{FB15k} \\
        \cmidrule(lr){2-3} \cmidrule(lr){5-6}
                                & \MRank & \HITten && \MRank & \HITten \\
        \midrule
        TransE~\citep{TransE}       
        & 251         & 89.2          && 125         & 47.1  \\
        TransH~\citep{TransH}       
        & 303         & 86.7          && 87          & 64.4 \\
        TransR~\citep{TransR}       
        & 225         & 92.0          && 77          & 68.7 \\
        CTransR~\citep{TransR}      
        & 218         & 92.3          && 75          & 70.2 \\
        KG2E~\citep{KG2E}         
        & 331         & 92.8          && \textbf{59} & 70.4 \\
        TransD~\citep{TransD}       
        & 212         & 92.2          && 91          & 77.3 \\
        TranSparse~\citep{TranSparse}   
        & \textbf{211}& 93.2          && 82          & 79.5 \\
        DistMult~\citep{DistMult}     
        & -           & 93.6          && -           & 82.4 \\
        ComplEx~\citep{ComplEx}      
        & -           & 94.7          && -           & \textbf{84.0} \\
        HolE~\citep{HolE}         
        & -           & \textbf{94.9} && -           & 73.9 \\
        \method      & 380          & \textbf{94.9} && 73          & 83.3 \\
        \bottomrule
    \end{tabular}
          \begin{tablenotes}
        \setlength\labelsep{0pt}
	\begin{footnotesize}
	\item
          The best performance under each metric is \textbf{bold}.
          \HITten values are in percent. 
          \par
	\end{footnotesize}
      \end{tablenotes}
  \end{threeparttable}
\end{table}
%

%
Table~\ref{tbl:results_wn18_fb15k} presents the comparison between \method and another
10 state-of-the-art 
methods on WN18 and FB15k datasets. The results of the other 10 methods are cited from
their respective publications. 
On WN18, \method achieves the best \HITten (94.9\%), tied with HolE. 
%
%
In terms of \MRank
\method is not as good as many of the others. This indicates that \method is able to rank the majority
positive testing triplets very high among top 10, but a small portion of the positive testing triplets
lower than other methods.
However, given the large number of entities in WN18 (i.e., \~41k as in Table~\ref{tbl:data}), 
we consider the performance of \method in terms of \MRank is
still reasonable (i.e., 1-380/40,943=99.07\%ile).  
On FB15k, \method is the 2nd best method in terms of \MRank (73), 
slightly worse than that of KG2E (\MRank = 59).
In terms of \HITten, \method is the 2nd best method (\HITten = 83.3\%),
only very slightly worse than ComplEx (\HITten = 84.0\%).
We believe \HITten is more meaningful in real applications when only a top few predictions
will be further investigated, and when it is very costly to validate every one in the top, for example,
211 predictions (TranSparse's \MRank = 211).
Thus, \method achieves the state-of-the-art performance in terms of \HITten on WN18 and FB15k, and
comparable performance on \MRank, compared with the other methods.  
Note that we also conducted experiments in which only the primary chain  in \method is used and
the secondary chain is removed. However, the performance of this simplified \method is 
worse than \method. For example, on FB15k, this method achieves \HITten = 79.0\%, compared to
that of \method (83.3\%). Thus, we did not present such results in the tables. 
Please also note that, although we present the experimental results on 
FB15k and WN18 as they are widely used benchmarks,
these two datasets may not properly evaluate the methods
due to their serious test leakage issue as reported in~\citep{ConvE}. 
In order to further evaluate our method, we conduct experiments on more difficult datasets, 
in which the interactions among triplets are more complicated.

\begin{table}[!h]	
  \caption{Entity Prediction on YAGO3-10 and FB15k-237}	
  \label{tbl:results_YAGO3_FB15K-237}	
  \centering
  \begin{threeparttable}
    \begin{tabular}{
	@{\hspace{10pt}}l@{\hspace{18pt}}
	@{\hspace{18pt}}r@{\hspace{18pt}}
	@{\hspace{18pt}}r@{\hspace{18pt}}
	@{\hspace{18pt}}c@{\hspace{18pt}}
        @{\hspace{18pt}}r@{\hspace{18pt}}
	@{\hspace{18pt}}r@{\hspace{10pt}}
      }
      \toprule
      \multirow{2}{*}{method} & \multicolumn{2}{c}{YAGO3-10} & & \multicolumn{2}{c}{FB15k-237} \\
      \cmidrule(lr){2-3} \cmidrule(lr){5-6}
                              & \MRank & \HITten && \MRank  & \HITten \\
      \midrule
      DistMult~\citep{DistMult} 
      & 5,926 & 54.0             && \textbf{254} & 41.9 \\
      ComplEx~\citep{ComplEx}  
      & 6,351 & 55.0             &&	        339 & 42.8 \\
      R-GCN~\citep{R-GCN}    
      &    - & -                &&    -  & 41.7 \\
      \method  & \textbf{2,061} & \textbf{62.9}    &&	        267 & \textbf{46.2} \\
      \bottomrule
    \end{tabular}	      
    \begin{tablenotes}
      \setlength\labelsep{0pt}
      \begin{footnotesize}
      \item
          The best performance under each metric is \textbf{bold}.
          \HITten values are in percent. \par
      \end{footnotesize}
    \end{tablenotes}
\end{threeparttable}
\end{table}	

%
Table~\ref{tbl:results_YAGO3_FB15K-237}	 presents the comparison between \method and 3
state-of-the-art methods on YAGO3-10 and FB15k-237 datasets. 
The results of the 3 methods are citepd from~\citep{ConvE}. 
Other methods do not have results reported on
these two datasets. 
On YAGO3-10, \method achieves the best \HITten (62.9\%) and \MRank among all the methods.
%
On FB15k-237, \method also achieves the best \HITten and \MRank. 
%
%
Thus, based on both Table~\ref{tbl:results_wn18_fb15k} and Table~\ref{tbl:results_YAGO3_FB15K-237},
\method performs better than the other comparison methods on average. 
The state-of-the-art performance and significant improvement on these difficult datasets 
properly demonstrate the strong ability of \method in modeling interactions among triplets.
%

\begin{table}[!h]	
  \caption{Entity Prediction on DDI and DGI}	
  \label{tbl:results_dgi_ddi}	  
  \centering	  
  \begin{threeparttable}
      \begin{tabular}{	 
	  @{\hspace{12pt}}l@{\hspace{18pt}}
	  @{\hspace{18pt}}r@{\hspace{18pt}}
	  @{\hspace{18pt}}r@{\hspace{18pt}}
	  @{\hspace{18pt}}c@{\hspace{18pt}}
          @{\hspace{18pt}}r@{\hspace{18pt}}
	  @{\hspace{18pt}}r@{\hspace{12pt}}
	}	
	\toprule
        \multirow{2}{*}{method} & \multicolumn{2}{c}{DGI} & & \multicolumn{2}{c}{DDI} \\
        \cmidrule(lr){2-3} \cmidrule(lr){5-6}
                                & \MRank & \HITten && \MRank & \HITten \\
        \midrule
        DistMult~\citep{DistMult}  
        & 157 & 86.6          && 66 & 22.0 \\
        ComplEx~\citep{ComplEx}   
        & 174 & \textbf{87.3} && 64 & 31.6 \\
        \method   &  \textbf{88}  & 86.8 && \textbf{35} & \textbf{35.7} \\
        \bottomrule	        
      \end{tabular}	      
      \begin{tablenotes}
        \setlength\labelsep{0pt}
	\begin{footnotesize}
	\item
          The best performance under each metric is \textbf{bold}.
          \HITten values are in percent. 
          \par
	\end{footnotesize}
      \end{tablenotes}
  \end{threeparttable}	
\end{table}	

%
Table~\ref{tbl:results_dgi_ddi} presents the comparison between \method, DistMult
and ComplEx.
%
On DGI, \method achieves the best \MRank (88) compared to DistMult and ComplEx,
and the 2nd best \HITten (86.8\%), which is slightly worse than ComplEx's.
On DDI, \method achieves the best \MRank and \HITten compared to DistMult and ComplEx,
and the improvement over ComplEx, the 2nd best method, is very significant (i.e., 45\% in \MRank
and 13\% in \HITten).

\subsection{Entity Prediction with Descriptions}

\subsubsection{Entity Prediction from Structure and Textual Information}

We use FB14k to evaluate the performance of \methodD. FB14k 
is a benchmark dataset to evaluate knowledge graph learning with descriptions~\citep{DKRL}.
The statistics of FB14k is presented in Table~\ref{tbl:data}.

In \methodD, we use Glove~\citep{Glove} pre-trained 100-dimension word vectors
to initialize the embeddings for each word in the entity descriptions.
For each entity, we use the first 200 words in its description  with necessary
padding, since 99.2\% entities have descriptions of fewer than 200 words. 
We compared \methodD with DKRL~\citep{DKRL} and SSP~\citep{SSP} as they
reported the state-of-the-art performance on FB14k.
For DKRL, when both the entity embeddings from structure information and from entity
descriptions are used in prediction, DKRL is denoted as DKRL (ALL). When only the
entity descriptions are used, the method is denoted as DKRL (CNN). 
For SSP, when the knowledge graph learning model and the topic model are trained separately, 
SSP is denoted as SSP (Std.).
When the tow models are trained simultaneously, it is denoted as SSP (Joint).

%
We conduct three sets of experiments on \methodD, 
following the experimental setting of Xie \etal~\citep{DKRL}.
In the first experiment, we use the mean of the \scoretxt and \scorestr as the predicted score,
as exactly in \methodD.  
In the second experiment, we only use \scorestr in \methodD, not \scoretxt, as the predicted score.
This method is denoted as $\methodD \text{(str)}$. 
In the third experiment, we only use \scoretxt in \methodD, not \scorestr, as the predicted score.
This method is denoted as $\methodD \text{(txt)}$.
Table~\ref{tbl:results_fb14k} presents the experimental results from all the comparison methods
on FB14k. 

\begin{table}[!h]
  \caption{Entity Prediction with Descriptions on FB14k}
  \label{tbl:results_fb14k}
  \centering
  \begin{threeparttable}
    \begin{tabular}{
	@{\hspace{8pt}}l@{\hspace{8pt}}
	@{\hspace{60pt}}r@{\hspace{60pt}}
	@{\hspace{8pt}}r@{\hspace{8pt}}
	}
      \toprule
      method & \MRank & \HITten \\
      \midrule
      DKRL (CNN)~\citep{DKRL} 
      & 113 & 57.6 \\
      DKRL (ALL)~\citep{DKRL} 
      & 91  & 67.4 \\
      SSP (Std.)~\citep{SSP} 
      & \textbf{77}  & 78.6 \\
      SSP (Joint)~\citep{SSP} 
      & 82 & 79.0 \\
      \method & 109 & 82.7 \\
      $\methodD {\text{(str)}}$ & 101 & 75.7\\
      $\methodD {\text{(txt)}}$ & 107 & 84.1 \\
      \methodD & 109 & \textbf{84.5} \\
      \bottomrule
      \end{tabular}
          \begin{tablenotes}
        \setlength\labelsep{0pt}
	\begin{footnotesize}
	\item
          $\methodD \text{(str)}$ and $\methodD \text{(txt)}$ represent that in \methodD,
          only \scorestr or \scoretxt is used for prediction, respectively. 
          The best performance under each metric is \textbf{bold}. 
          \HITten values are in percent. \par
	\end{footnotesize}
      \end{tablenotes}
  \end{threeparttable}
\end{table}
%

%
Table~\ref{tbl:results_fb14k} demonstrates that \methodD achieves the
best \HITten (84.5\%) among all the methods. $\methodD \text{(txt)}$ achieves
the second best \HITten (84.1\%) and $\methodD \text{(str)}$ achieves a poor
\HITten (75.7\%), even worse than that from \method (82.7\%) when no textual
information is used.
Both $\methodD \text{(txt)}$ and $\methodD \text{(str)}$ train a \methodD
model but use different scores for prediction. The good performance of
$\methodD \text{(txt)}$ indicates that textual information is useful in
entity prediction, while the worse performance of $\methodD \text{(str)}$
than \method demonstrates the effectiveness of the dual-chain architecture
in \method (i.e., only structure information is used in the model)
in improving performance and reducing overfitting.
The fact that \methodD outperforms both $\methodD \text{(txt)}$ and $\methodD \text{(str)}$
indicates that the structure and textual information might be complementary and
combining them will benefit entity prediction.
The fact that \methodD outperforms \method indicates the capability of \methodD
in integrating structure information and textual information for entity prediction.

\subsubsection{Zero-Shot Learning}
%

Most traditional knowledge graph methods can not handle new entities that are
not involved in the knowledge graph during training. 
Our \methodD provides a natural way to deal with this zero-shot problem by learning
entity representations through their textual descriptions, and using the scores predicted
from the textual information (i.e., \scorestr in \methodD) for prediction (i.e., $\methodD \text{(txt)}$
as in Table~\ref{tbl:results_fb14k}). 
We compare \methodD with another two methods CBOW~\citep{DKRL} and DKRL~\citep{DKRL}, which are able to
deal with zero-shot learning. We test these methods on dataset FB20k~\citep{DKRL}, in which
each testing triplet has at least one new entity.
In our experiments, the testing set is split into 3 subsets with
different types of triplets: 1). triplets with new entities at head, denoted as \newhead; 2). triplets
with new entities at tail, denoted as \newtail; and 3). triplets with new entities at both heads and tails,
denoted as \newht. Table~\ref{tbl:data_fb20} presents the FB20K data statistics on \newhead, \newtail and \newht. 
\begin{table}[!h]
  \caption{Statistics of FB20k}
  \label{tbl:data_fb20}
  \centering
  \begin{threeparttable}
      \begin{tabular}{
	@{\hspace{5pt}}l@{\hspace{5pt}}
	@{\hspace{10pt}}r@{\hspace{10pt}}          
	@{\hspace{10pt}}r@{\hspace{10pt}}
    @{\hspace{10pt}}r@{\hspace{10pt}}
    @{\hspace{10pt}}r@{\hspace{10pt}}
    @{\hspace{10pt}}r@{\hspace{10pt}}
	}
        \toprule
        dataset  & \#entity & \#\newhead & \#\newtail & \#\newht &\#total\\
        \midrule
        FB20k & 19,923 & 18,753 & 11,586 & 151 & 30,490\\
        \bottomrule
      \end{tabular}
      \begin{tablenotes}
        \setlength\labelsep{0pt}
	\begin{footnotesize}
	\end{footnotesize}
      \end{tablenotes}
  \end{threeparttable}
\end{table}
%


%
We test \methodD, CBOW and DKRL on these three subsets and present the results in
Table~\ref{tbl:results_zero_shot}.
We use the \methodD model that is trained from FB14k and performs best on FB14k
to test FB20k, following the setting in~\citep{DKRL}.  
Our results demonstrate that $\methodD \text{(txt)}$
substantially outperforms CBOW and DKRL on \newhead and \newtail.
In particular, $\methodD \text{(txt)}$ outperforms DKRL (the second best method)
30.7\% and 40.2\% on \newhead and \newtail, respectively. 
On \newht, $\methodD \text{(txt)}$ is slightly worse than DKRL (CNN) but still 3.9\% better
than CBOW.
However, given the very limited size of \newht, 
the slight difference between $\methodD \text{(txt)}$ and
DKRL (CNN) (i.e., only 4 missed hits in $\methodD \text{(txt)}$) is not significant.
Overall, $\methodD \text{(txt)}$ is 33.2\%
better than DKRL (CNN)
in terms of the weighted-sum performance over \newhead, \newtail and \newht together
(i.e., total in Table~\ref{tbl:results_zero_shot}). 
These results demonstrate the strong capability of \methodD in dealing with zero-shot 
problems in knowledge graph learning. 
\begin{table}[!h]
  \caption{Zero-Shot Learning on FB20k (\HITten)}
  \label{tbl:results_zero_shot}
  \centering
  \begin{threeparttable}
    \begin{tabular}{
	@{\hspace{12pt}}l@{\hspace{18pt}}
	@{\hspace{18pt}}r@{\hspace{18pt}}          
	@{\hspace{18pt}}r@{\hspace{18pt}}
        @{\hspace{18pt}}r@{\hspace{18pt}}
        @{\hspace{18pt}}r@{\hspace{12pt}}
	}
        \toprule
        method & \newhead & \newtail & \newht & total \\
        \midrule

        CBOW~\citep{SSP}                    
        & 27.1 & 21.7 & 67.2 & 24.6 \\
        DKRL (CNN)~\citep{SSP}              
        & 31.2 & 26.1 & \textbf{72.5} & 29.5 \\
        $\methodD \text{(txt)}$ & \textbf{40.8} & \textbf{36.6} & 69.8 & \textbf{39.3} \\
        \bottomrule
      \end{tabular}
          \begin{tablenotes}
        \setlength\labelsep{0pt}
	\begin{footnotesize}
	\item
          The column of ``total'' has the weighted sum of \HITten
          over the three cases \newhead, \newtail and \newht. 
          The best performance under each metric is \textbf{bold}.
                    \HITten values are in percent. 
          \par
	\end{footnotesize}
          \end{tablenotes}
  \end{threeparttable}
\end{table}
%


\subsection{Representation Distribution}

\begin{figure}[!htbp]
  \centering
  \input{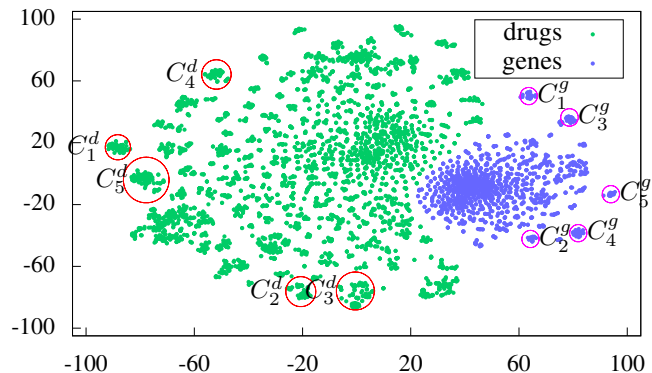}
  \caption{Embeddings from \method on DGI}
  \label{fig:embed_CNN}
  \vspace{-10pt}
\end{figure}
\begin{figure}[!htbp]
  \centering
  \input{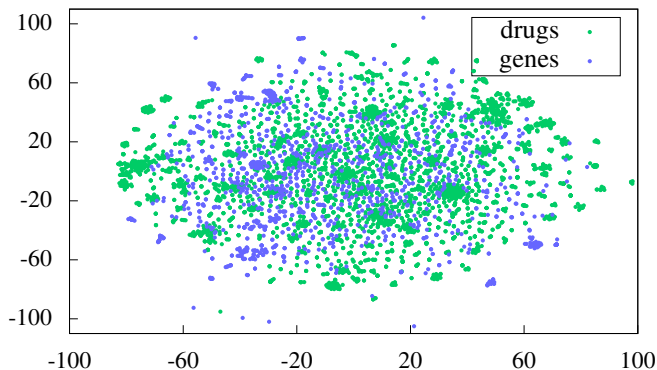}
  \caption{Embeddings from ComplEx (Real Vectors) on DGI}
  \label{fig:embed_Complex_real}
  \vspace{-10pt}
\end{figure}
We compare the embeddings learned from \method and ComplEx on the DGI dataset,
as ComplEx is very strong baseline according to Table~\ref{tbl:results_dgi_ddi}.
We use t-SNE method \citep{maaten2008visualizing} to project the embeddings into 2-dimension
space. Figure~\ref{fig:embed_CNN} presents the embeddings of drugs and genes learned from \method
in 2D space. Figure~\ref{fig:embed_Complex_real} presents the embeddings of the real vectors
of drug and genes learned from ComplEx.
The projected embeddings of 
imaginary vectors are very similar to that in Figure~\ref{fig:embed_Complex_real} so we don't present
them here. 
Please note that in these two figures, a small Euclidean distance between a drug and a gene does not 
necessarily mean that the drug and gene tend to interact due to the ways the interactions are
calculated in \method (i.e., through CNN) and in ComplEx (i.e., through bilinear scoring).  


In Figure~\ref{fig:embed_CNN}, the drugs and genes are well separated by \method, and there
are some tight clusters among drugs and genes, respectively. However, in Figure~\ref{fig:embed_Complex_real},
the drugs and genes are all mixed, and the clustering structure is less clear. 
We checked several clusters in Figure~\ref{fig:embed_CNN}.
Most of the drugs in drug cluster 1
in Figure~\ref{fig:embed_CNN} (i.e., $C_1^d$) are anti-inflammatory drugs
(e.g., betamethasone, dexamethasone, fluorometholone, hydrocortamate and prednisolone),
and belong to drug class corticosteroid and glucocorticoid.
Most of the drugs in drug cluster 2 (i.e., $C_2^d$) are used to treat anxiety, tension, depression and insomnia
(e.g., diazepam, meprobamate, methyprylon, temazepam and zaleplon). 
Most of the drugs in drug cluster 5 (i.e., $C_5^d$) are used to treat allergy symptoms and allergic rhinitis
(e.g., azelastine, cetirizine, clemastine, dimethindene and triprolidine). 
%
%
There are also some tight gene clusters. For example, in gene cluster 1 (i.e., $C_1^g$) in
Figure~\ref{fig:embed_CNN}, almost all the genes are from gene group NADH:ubiquinone oxidoreductase core
subunits (NDUF, MT-ND) (e.g., MT-ND5, MT-ND6, NDUFB9, NDUFV1 and NDUFS4).  
In gene cluster 5 (i.e., $C_5^g$), almost all the genes are from gene group Potassium voltage-gated channels (e.g., KCNQ5,
KCNH6, KCNA2).
Since drugs from a same drug class tend to be similar in terms of their chemical
structures and/or their interactions with protein pathways, and genes from a same gene
family tend to interact with similar drugs, 
the clustering structures among the drug and gene embeddings from \method demonstrate that
\method is able to learn the entity relations that conform to knowledge.  


\subsection{Case Study}

\begin{table}[!h]
  \caption{Examples of \method Prediction on DDI Dataset}
  \label{tbl:ddi_case}
  \centering
  \begin{threeparttable}
    \begin{tabular}{	 
    	  @{\hspace{0pt}}l@{\hspace{12pt}}
	  @{\hspace{12pt}}l@{\hspace{12pt}}
          @{\hspace{12pt}}l@{\hspace{12pt}} 
	  @{\hspace{12pt}}l@{\hspace{0pt}}
 	}	
	\toprule
        & drug 1 & drug 2 & side effects \\
        \midrule
        \multirow{5}{*}{\rotatebox[origin=c]{90}{\parbox[c]{1.4cm}{\centering head prediction}}}
        & fluconazole & metformin     & enuresis \\
        & guaifenesin & telithromycin & respiratory alkalosis  \\
        & amlodipine  & sumatriptan   & breast disorder  \\
        & voltaren    & a-methapred & cystitis interstitial  \\
        & furosemide  & rofecoxib & ingrowing nail\\
        \midrule
        \multirow{5}{*}{\rotatebox[origin=c]{90}{\parbox[c]{1.4cm}{\centering tail prediction}}}
        & bupropion    & acetaminophen & keratoconjunctivitis sicca  \\
        & cyclobenzaprine & ibuprofen  & epicondylitis \\
        & amitriptyline & carbamazepine & trigger finger \\
        & clonazepam & gabapentin & enuresis  \\
        & diazepam  & primidone & bunion\\
        \bottomrule
    \end{tabular}
  \end{threeparttable}
\end{table}
%

%
Table~\ref{tbl:ddi_case} presents some examples from head prediction (\newhead)
and tail prediction (\newtail) that \method was able to correctly
rank among top 10 in the corresponding testing lists, but ComplEx was not. 
The examples represented in Table~\ref{tbl:ddi_case} do not
involve either rare drugs or rare side effects. Actually, they correspond to
average drug and side effects in the dataset.
Thus, Table~\ref{tbl:ddi_case} shows that \method is able to better prioritize
true relations compared to ComplEx. 

\section{Conclusion and Future Work}
\label{sec:conclusion}

In this paper, we developed a CNN-based dual-chain model (\method) for
knowledge graph learning. In \method, two learning chains are constructed,
one learning from the (\head, \relation, \tail) triplet embeddings via a CNN and the
following layers, the other learning from a sparsified version of the triplet
embeddings via exactly same architectures as in the first chain.
Our experimental results demonstrate that this CNN based method for triplet
embedding learning, together with the dual-chain structure, is able to achieve
or outperform the state-of-the-art results on graph learning.
We also extended \method to incorporate textual information (entity descriptions)
into \methodD. Embeddings from descriptions are also learned via CNN and structure attention
for the entities, in addition to the embeddings learned from knowledge graph structures.
The two types of embeddings are both used to predict entities in \methodD.
Therefore, \methodD is able to work for new entities that are not available during
model training (i.e., zero-shot problem). 
Our experimental results demonstrate that \methodD outperforms other methods 
that are also able to handle zero-shot problems.
In the future research, we will also incorporate the path and community information from
the knowledge graph structures, in addition to textual information on entities, to further improve
learning performance. We will also consider using the textual information of the entities
along a path or within a community.
We will also try and apply such methods in more impactful applications such as disease-gene
association graph, etc.

\bibliographystyle{spbasic}
\bibliography{paper}

\begin{thebibliography}{25}
\providecommand{\natexlab}[1]{#1}
\providecommand{\url}[1]{{#1}}
\providecommand{\urlprefix}{URL }
\expandafter\ifx\csname urlstyle\endcsname\relax
  \providecommand{\doi}[1]{DOI~\discretionary{}{}{}#1}\else
  \providecommand{\doi}{DOI~\discretionary{}{}{}\begingroup
  \urlstyle{rm}\Url}\fi
\providecommand{\eprint}[2][]{\url{#2}}

\bibitem[{Bollacker et~al.(2008)Bollacker, Evans, Paritosh, Sturge, and
  Taylor}]{bollacker2008freebase}
Bollacker K, Evans C, Paritosh P, Sturge T, Taylor J (2008) Freebase: a
  collaboratively created graph database for structuring human knowledge. In:
  Proceedings of the 2008 ACM SIGMOD international conference on Management of
  data, AcM, pp 1247--1250

\bibitem[{Bordes et~al.(2013)Bordes, Usunier, Garcia-Duran, Weston, and
  Yakhnenko}]{TransE}
Bordes A, Usunier N, Garcia-Duran A, Weston J, Yakhnenko O (2013) Translating
  embeddings for modeling multi-relational data. In: Advances in neural
  information processing systems, pp 2787--2795

\bibitem[{Cotto et~al.(2017)Cotto, Wagner, Feng, Kiwala, Coffman, Spies,
  Wollam, Spies, Griffith, and Griffith}]{DGIdb}
Cotto KC, Wagner AH, Feng YY, Kiwala S, Coffman AC, Spies G, Wollam A, Spies
  NC, Griffith OL, Griffith M (2017) Dgidb 3.0: a redesign and expansion of the
  drug--gene interaction database. Nucleic acids research 46(D1):D1068--D1073

\bibitem[{Dettmers et~al.(2018)Dettmers, Minervini, Stenetorp, and
  Riedel}]{ConvE}
Dettmers T, Minervini P, Stenetorp P, Riedel S (2018) Convolutional 2d
  knowledge graph embeddings. In: Proceedings of the Thirty-Second {AAAI}
  Conference on Artificial Intelligence, (AAAI-18), the 30th innovative
  Applications of Artificial Intelligence (IAAI-18), and the 8th {AAAI}
  Symposium on Educational Advances in Artificial Intelligence (EAAI-18), New
  Orleans, Louisiana, USA, February 2-7, 2018, pp 1811--1818

\bibitem[{Glorot and Bengio(2010)}]{glorot2010understanding}
Glorot X, Bengio Y (2010) Understanding the difficulty of training deep
  feedforward neural networks. In: Proceedings of the thirteenth international
  conference on artificial intelligence and statistics, pp 249--256

\bibitem[{He et~al.(2015)He, Liu, Ji, and Zhao}]{KG2E}
He S, Liu K, Ji G, Zhao J (2015) Learning to represent knowledge graphs with
  gaussian embedding. In: Proceedings of the 24th ACM International on
  Conference on Information and Knowledge Management, ACM, pp 623--632

\bibitem[{Ji et~al.(2015)Ji, He, Xu, Liu, and Zhao}]{TransD}
Ji G, He S, Xu L, Liu K, Zhao J (2015) Knowledge graph embedding via dynamic
  mapping matrix. In: Proceedings of the 53rd Annual Meeting of the Association
  for Computational Linguistics and the 7th International Joint Conference on
  Natural Language Processing (Volume 1: Long Papers), vol~1, pp 687--696

\bibitem[{Ji et~al.(2016)Ji, Liu, He, and Zhao}]{TranSparse}
Ji G, Liu K, He S, Zhao J (2016) Knowledge graph completion with adaptive
  sparse transfer matrix. In: Proceedings of the Thirtieth {AAAI} Conference on
  Artificial Intelligence, February 12-17, 2016, Phoenix, Arizona, {USA.}, pp
  985--991

\bibitem[{Lacroix et~al.(2018)Lacroix, Usunier, and
  Obozinski}]{lacroix2018canonical}
Lacroix T, Usunier N, Obozinski G (2018) Canonical tensor decomposition for
  knowledge base completion. arXiv preprint arXiv:180607297

\bibitem[{Lehmann et~al.(2015)Lehmann, Isele, Jakob, Jentzsch, Kontokostas,
  Mendes, Hellmann, Morsey, Van~Kleef, Auer et~al.}]{dbpedia}
Lehmann J, Isele R, Jakob M, Jentzsch A, Kontokostas D, Mendes PN, Hellmann S,
  Morsey M, Van~Kleef P, Auer S, et~al. (2015) Dbpedia--a large-scale,
  multilingual knowledge base extracted from wikipedia. Semantic Web
  6(2):167--195

\bibitem[{Lin et~al.(2015)Lin, Liu, Sun, Liu, and Zhu}]{TransR}
Lin Y, Liu Z, Sun M, Liu Y, Zhu X (2015) Learning entity and relation
  embeddings for knowledge graph completion. In: Proceedings of the
  Twenty-Ninth {AAAI} Conference on Artificial Intelligence, January 25-30,
  2015, Austin, Texas, {USA.}, pp 2181--2187

\bibitem[{Lin et~al.(2017)Lin, Feng, dos Santos, Yu, Xiang, Zhou, and
  Bengio}]{StructrueAtt}
Lin Z, Feng M, dos Santos CN, Yu M, Xiang B, Zhou B, Bengio Y (2017) A
  structured self-attentive sentence embedding. CoRR abs/1703.03130

\bibitem[{Maaten and Hinton(2008)}]{maaten2008visualizing}
Maaten Lvd, Hinton G (2008) Visualizing data using t-sne. Journal of machine
  learning research 9(Nov):2579--2605

\bibitem[{Mahdisoltani et~al.(2015)Mahdisoltani, Biega, and Suchanek}]{YAGO3}
Mahdisoltani F, Biega J, Suchanek FM (2015) {YAGO3:} {A} knowledge base from
  multilingual wikipedias. In: {CIDR} 2015, Seventh Biennial Conference on
  Innovative Data Systems Research, Asilomar, CA, USA, January 4-7, 2015,
  Online Proceedings

\bibitem[{Miller et~al.(1990)Miller, Beckwith, Fellbaum, Gross, and
  Miller}]{wordnet}
Miller GA, Beckwith R, Fellbaum C, Gross D, Miller KJ (1990) Introduction to
  wordnet: An on-line lexical database. International journal of lexicography
  3(4):235--244

\bibitem[{Nickel et~al.(2016)Nickel, Rosasco, and Poggio}]{HolE}
Nickel M, Rosasco L, Poggio TA (2016) Holographic embeddings of knowledge
  graphs. In: Proceedings of the Thirtieth {AAAI} Conference on Artificial
  Intelligence, February 12-17, 2016, Phoenix, Arizona, {USA.}, pp 1955--1961

\bibitem[{Niepert et~al.(2016)Niepert, Ahmed, and Kutzkov}]{GCN}
Niepert M, Ahmed M, Kutzkov K (2016) Learning convolutional neural networks for
  graphs. In: International conference on machine learning, pp 2014--2023

\bibitem[{Pennington et~al.(2014)Pennington, Socher, and Manning}]{Glove}
Pennington J, Socher R, Manning C (2014) Glove: Global vectors for word
  representation. In: Proceedings of the 2014 conference on empirical methods
  in natural language processing (EMNLP), pp 1532--1543

\bibitem[{Schlichtkrull et~al.(2018)Schlichtkrull, Kipf, Bloem, van~den Berg,
  Titov, and Welling}]{R-GCN}
Schlichtkrull M, Kipf TN, Bloem P, van~den Berg R, Titov I, Welling M (2018)
  Modeling relational data with graph convolutional networks. In: European
  Semantic Web Conference, Springer, pp 593--607

\bibitem[{Tatonetti et~al.(2012)Tatonetti, Patrick, Daneshjou, and
  Altman}]{twosides}
Tatonetti NP, Patrick PY, Daneshjou R, Altman RB (2012) Data-driven prediction
  of drug effects and interactions. Science translational medicine
  4(125):125ra31--125ra31

\bibitem[{Trouillon et~al.(2016)Trouillon, Welbl, Riedel, Gaussier, and
  Bouchard}]{ComplEx}
Trouillon T, Welbl J, Riedel S, Gaussier {\'E}, Bouchard G (2016) Complex
  embeddings for simple link prediction. In: International Conference on
  Machine Learning, pp 2071--2080

\bibitem[{Wang et~al.(2014)Wang, Zhang, Feng, and Chen}]{TransH}
Wang Z, Zhang J, Feng J, Chen Z (2014) Knowledge graph embedding by translating
  on hyperplanes. In: Proceedings of the Twenty-Eighth {AAAI} Conference on
  Artificial Intelligence, July 27 -31, 2014, Qu{\'{e}}bec City, Qu{\'{e}}bec,
  Canada., pp 1112--1119

\bibitem[{Xiao et~al.(2017)Xiao, Huang, Meng, and Zhu}]{SSP}
Xiao H, Huang M, Meng L, Zhu X (2017) Ssp: Semantic space projection for
  knowledge graph embedding with text descriptions. In: AAAI, vol~17, pp
  3104--3110

\bibitem[{Xie et~al.(2016)Xie, Liu, Jia, Luan, and Sun}]{DKRL}
Xie R, Liu Z, Jia J, Luan H, Sun M (2016) Representation learning of knowledge
  graphs with entity descriptions. In: Proceedings of the Thirtieth {AAAI}
  Conference on Artificial Intelligence, February 12-17, 2016, Phoenix,
  Arizona, {USA.}, pp 2659--2665

\bibitem[{Yang et~al.(2014)Yang, Yih, He, Gao, and Deng}]{DistMult}
Yang B, Yih W, He X, Gao J, Deng L (2014) Embedding entities and relations for
  learning and inference in knowledge bases. CoRR abs/1412.6575

\end{thebibliography}

\newpage

\appendix
\section{Appendix}
\label{sec:appendix}

\subsection{Data Augmentation and Reproducibility Issues with ConvE}
\label{sec:comp:repro}

The training and testing data used in the ConvE are constructed in a different way
compared to the data used in \method and also the data used in
other baseline methods such as TransE, Distmult and Complex.
In ConvE, before training the model, for each triplet with reversible relation
$(\head, \relation_{rv}, \tail)$,
it adds a reversed triplet $(\tail, \relation_{rv}, \head)$ into the
training, validation and testing sets, correspondingly.
As a result, ConvE has benefited from this data augmentation
to achieve better performance~\citep{lacroix2018canonical}.
Moreover, the reversed triplets on the testing set
give the model two chances to predict a same triplet correctly. 
Thus, it further over-estimate the performance.
However, in our experiments and the experiments of the baseline methods,
such data augmentation before training is not implemented.
So it is not fair to compare the reported results of ConvE with the experimental
results of our model and baseline methods.
This is an issue that has been recognized by others~\footnote{\url{https://github.com/TimDettmers/ConvE/issues/45}}.
It is noteworthy that, without data augmentation,
\method can still achieve similar or even better performance
than ConvE on almost all the public datasets used in our experiments.
It demonstrates that the CNN architecture used in our model is very competitive
or even better compared to the CNN architecture used in ConvE.

We also find that the source code of ConvE published on Github~\footnote{\url{https://github.com/TimDettmers/ConvE}}
will cause memory explosion issues and thus
we can not reproduce the reported results in ConvE.
Our computers with 16GB RAM can not run their code through on FB15k-237, which is a small dataset
used in their paper, while it only takes about 3.5GB to run our model.
Because of this, we reimplement ConvE with Tensorflow and train it on FB15k.
However, tt produces much worse results compared with the results reported by ConvE.
Consequently upon these facts, we don't use ConvE as baseline in our experiments.


\subsection{Experimental Setting}
\label{sec:exp:setting}

%
We implemented the algorithm in python with tensorflow (www.tensorflow.org).

On the WN18, FB15k, FB15k-237 and DDI datasets, the learning rate starts from 3e-3;
on the YAGO3-10, FB14k and DGI datasets, the learning rate starts from 1e-3.
We decreased the learning rate in each epoch with a rate 0.998 on all the datasets.
%
%
We find the following hyper-parameters work well on the datasets:
on dataset WN18, the dimension of entity and relationship embedding $k$ as 200,
the number of convolutional kernel $n_c$ as 64,
the dimension of the triplet representations, denoted as $d_g$, as 256,
the drop-out rate $p$ as 0.2, batch size as 5,000, the number of training epoches as
3,000, and label smoothing weight as 0.0.
On dataset FB15k, the best performing parameters are as follows: $k$=200,
$n_c$=64,
$d_g$=256,
$p$=0.2, batch size=5,000, number of epochs=3,000 and label smoothing weight=0.0;
On dataset FB15k-237, the best performing parameters are as follows: $k$=200,
$n_c$=64,
$d_g$=256,
$p$=0.2, batch size=2,000, number of epochs=2,000 and label smoothing weight=0.1.
On dataset DDI, the best performing parameters are as follows: $k$=200,
$n_c$=64,
$d_g$=256,
$p$=0.2, batch size=300, number of epochs=1,000 and label smoothing weight=0.0.
On dataset YAGO3-10, the best performing parameters are as follows: $k$=200,
$n_c$=64,
$d_g$=256,
$p$=0.2, batch size=30,000, number of epochs=2,000 and label smoothing weight=0.0.
On dataset FB14k, the best performing parameters are as follows: $k$=100,
$n_c$=64,
$d_g$=256,
$p$=0.2, batch size=1,000, number of epochs=600 and label smoothing weight=0.0.
On dataset DDI, the best performing parameters are as follows: $k$=200,
$n_c$=64,
$d_g$=256,
$p$=0.2, batch size=100, number of epochs=1,000 and label smoothing weight=0.0.
In order to avoid overfitting, we regularized the embedding of relations
and entities to have unit $\ell_2$ norm.
We will publish the source code upon the acceptance of this paper.

\subsection{Comparison between \method and ConvE on experimental results}
Tabel~\ref{tbl:compare_with_ConvE} presents the comparison between \method and ConvE.
The results of ConvE are cited from its paper.
On almost all the datasets, without data augmentation,
\method can still achieve similar or even better performance than ConvE.
It demonstrates that the CNN architecture used in
\method is very competitive or even better compared to the CNN architecture used in ConvE.

\begin{table}[!h]       
  \caption{Comparison between \method and ConvE}
  \label{tbl:compare_with_ConvE}   
  \centering
  \begin{threeparttable}
    \begin{tabular}{
        @{\hspace{3pt}}l@{\hspace{8pt}}
        @{\hspace{8pt}}r@{\hspace{8pt}}
        @{\hspace{8pt}}r@{\hspace{8pt}}
        @{\hspace{4pt}}c@{\hspace{4pt}}
        @{\hspace{8pt}}r@{\hspace{8pt}}
        @{\hspace{8pt}}r@{\hspace{8pt}}
        @{\hspace{4pt}}c@{\hspace{4pt}}
        @{\hspace{8pt}}r@{\hspace{8pt}}
        @{\hspace{8pt}}r@{\hspace{8pt}}
        @{\hspace{4pt}}c@{\hspace{4pt}}
        @{\hspace{8pt}}r@{\hspace{8pt}}
        @{\hspace{8pt}}r@{\hspace{3pt}}
      }
      \toprule
      \multirow{2}{*}{method} & \multicolumn{2}{c}{WN18} & & \multicolumn{2}{c}{FB15k} & & \multicolumn{2}{c}{YAGO3-10} & & \multicolumn{2}{c}{FB15k-237} \\
      \cmidrule(lr){2-3} \cmidrule(lr){5-6} \cmidrule(lr){8-9} \cmidrule(lr){11-12}
                              & \MRank & \HITten && \MRank  & \HITten && \MRank & \HITten && \MRank  & \HITten\\
      \midrule
      ConvE & {374} & {95.6} && {51} & 83.1 && {1,676} & 62.0 &&  {244}  & {50.1}    \\
      \method  & 380 & 94.9 && 73 & {83.3} && 2,061 & {62.9} && 267 & 46.2 \\
      \bottomrule
    \end{tabular}             
    \begin{tablenotes}
      \setlength\labelsep{0pt}
      \begin{footnotesize}
      \item
          \HITten values are in percent.
        \par
      \end{footnotesize}
    \end{tablenotes}
\end{threeparttable}
\end{table}



\end{document}